\documentclass[12pt]{article}

\usepackage{latexsym}
\usepackage{amssymb}
\usepackage{amsmath}
\usepackage[longnamesfirst]{natbib}
\usepackage{graphicx}
\usepackage{fancyhdr}
\usepackage{bbm}

\usepackage{lineno}

\newcommand{\TIT}{{\it Threshold Martingale \tiny{(\today)}}}

\newcommand{\yhat}{{\widehat{Y}}}
\newcommand{\ytilde}{{\widetilde{Y}}}

\newcounter{remark}
\newcommand{\remark}[1]{\addtocounter{remark}{1} {\vspace*{0.1in}}\noindent {\it Remark \Alph{remark}.  }{#1}\vspace*{0.2in}}

\newcommand{\half}{\mbox{$\frac{1}{2}$}}

\newcommand{\smfrac}[2]{\mbox{$\frac{\scriptstyle #1}{\scriptstyle #2}$}}

\newcommand{\iid}{\stackrel {\mbox{\scriptsize iid}}{\sim}}
\newcommand{\pr}{\mathbb{P}\,}
\newcommand{\ev}{\mathbb{E}}

\newcommand{\R}{\mbox{$\mathbb{R}$}}

\newcommand{\tr}{\mbox{tr}}

\newcommand{\one}{{\mathbbm{1}}}

\newcommand{\ie}{\mbox{\it i.e.}}
\newcommand{\eg}{\mbox{\it e.g.}}

\newtheorem{theorem}{Theorem}

\newcommand{\be}{\begin{equation*}}
\newcommand{\qe}{\end{equation*}}

\newcommand{\ble}{\begin{equation}}
\newcommand{\qle}[1]{\label{eq:#1}\end{equation}}

\newcommand{\ip}[2]{\mbox{$\langle #1,\, #2 \rangle$}}
\newcommand{\norm}[1]{\mbox{$\| #1 \|$}}

\newcommand{\Cov}{\,\mbox{Cov}}

\newcommand{\Var}{\,\mbox{Var}}

\newcommand{\al}{\alpha}

\newcommand{\ep}{\epsilon}

\newcommand{\eqn}[1]{(\ref{eq:#1})}

\newcommand{\given}{\mid}
\newcommand{\QED}{\frame{\rule{0pt}{6pt}\rule{6pt}{0pt}}}

\newcommand{\ol}[1]{\overline{#1}}

\fancypagestyle{firstpage}{%
  \lhead{}
  \rhead{\tiny \today}
  
  \rfoot{}
}

\title{  
\vspace{-0.5em}
        Threshold Martingales and the Evolution of Forecasts
\vspace{-0.5em}
}

\author{
        Dean P. Foster and Robert A. Stine\footnote{All correspondence
regarding this manuscript should be directed to Prof. Stine at 
the address shown with the title.  He can be reached via e-mail at
stine@wharton.upenn.edu.}                                    \\
        Department of Statistics                                  \\
        The Wharton School of the University of Pennsylvania \\
        Philadelphia, PA 19104-6340                          \\
}

\date{\today}

\begin{document}
\maketitle

\abstract{

This paper introduces a martingale that characterizes two properties of evolving forecast distributions. Ideal forecasts of a future event behave as martingales, sequentially updating the forecast to leverage the available information as the future event approaches. The threshold martingale introduced here measures the proportion of the forecast distribution lying below a threshold. In addition to being calibrated, a threshold martingale has quadratic variation that accumulates to a total determined by a quantile of the initial forecast distribution.  Deviations from calibration or total volatility signal problems in the underlying model.  Calibration adjustments are well-known, and we augment these by introducing a martingale filter that improves volatility while guaranteeing smaller mean squared error.  Thus, post-processing can rectify problems with calibration and volatility without revisiting the original forecasting model. We apply threshold martingales first to forecasts from simulated models and then to models that predict the winner in professional basketball games.
 } %

\pagebreak

\section{ Introduction }

We study the evolution of probabilistic forecasts of a future event that specify a distribution. The distributions of forecasts evolve in the sense that they change as the passage of time reveals more information about the approaching event.  For example, we could be predicting demand for holiday gifts. Planning for this demand using a model such as the classic news-vendor requires more than a point estimate; one needs a distribution.  The forecast distribution would be highly uncertain months ahead of the holidays, and then become increasingly precise as we learn more regarding the impact of economic conditions on consumer spending.  Or, imagine estimating the chances that the home team wins a basketball game.  A lead of 5 points early in the game doesn't reveal nearly so much about the chances of a home win as a lead of the same size near the end of the game.  In both cases, probabilities change as target event approaches.  Our interest here is quantifying whether the forecasts of the evolve as they should, using a martingale to define the gold standard.

The use of martingales to characterize evolving forecasts is not novel.  For example, \citet{heath94} propose the martingale model of forecast evolution (MMFE) as a means to simulate forecast systems. Suppose that we observe a time series up through time $t$, say $Y_1, \, Y_2, \ldots, Y_t$, and want to predict the series at some future time $T$.  Everything done in this paper applies to vector time series $Y_t \in {\mathbb R}^d$. We stick to the scalar case $d=1$ for clarity and to keep the notation simple.  We refer to $T$ as the forecast target date (FTD) and denote the forecast of $Y_T$ created at time $t < T$ by $\yhat_{T|t}$.  One expects the magnitude of the forecast error $Y_T - \yhat_{T|t}$ to get smaller as $t$ approaches $T$, but the MMFE allows one to say more.  Martingales capture the notion that the forecast $\yhat_{T|t}$ should capture {\em all} of the information concerning $Y_T$ that is available at time $t$. If forecasts $\yhat_{T|t}$ meet the conditions of the MMFE, one can characterize how the forecasts $\yhat_{T|t}$ change, or evolve, as the historical data approach the FTD.  Furthermore, the MMFE does not require one to know all the details of the forecasting system; the forecast might be the result of a neural network or a heuristic spreadsheet procedure. All that is required is that forecasts conform to the requirements of a martingale. Given that, one can estimate means and variances from data.  These estimates allow one to simulate the forecasting system.

Our use of martingales differs from the approach adopted in the MMFE. We emphasize diagnostic methods intended to check if the forecasts evolve appropriately.  In addition, rather than examine $\yhat_{T|t}$ directly, we focus on a sequence of probabilities defined by quantiles of the evolving forecast distributions. If the sequence of forecasts is a martingale, then so too are these probabilities.  As a result, the threshold probabilities should be calibrated and exhibit a known level of volatility.  Deviations from these indicate flaws in the forecasts which can be corrected by post processing.

The remainder of this paper develops as follows.  The following section briefly reviews discrete martingales and their connection to familiar autoregressive models of stationary time series, keeping this paper more self-contained.  Section 3 then defines the threshold martingales that define our diagnostic procedure.  The following two sections give examples, first simulating properties under known models (Section 4) and then applying the method to basketball scoring (Section 5). The paper concludes with a brief discussion of extensions.  An appendix defines the martingale filter that we use in Section 5 to reduce volatility.

\section{ Martingales }

This section reviews the definition of a martingale and connects it to forecasting, autoregressive processes, and other statistical concepts.  Though martingales are likely familiar to many readers as a tool for understanding probability theory, we emphasize the connections between martingales and statistics and show how martingales are related to the evolution of forecasts.  

In discrete time, a martingale is a stochastic process $\{X_t\}$ for which the conditional expected value of $X_{t+1}$ given its predecessors is the most recent value,
\ble
    \ev (X_{t+1} \given X_{t},\, X_{t-1},\, \ldots,\,X_0) = \ev(X_{t+1} \given {\cal F}_t) = X_t
\qle{evx}
The sigma field ${\cal F}_t$ collects all of the information available from $X_t, \, X_{t-1}, \ldots,\,X_0$.\footnote{ In general, the sequence of nested sigma fields $\{{\cal F}_t\}$ (a filtration) in the definition of a martingale represents {\em all} information available at time $t$, not just that represented by prior random variables.  For our purposes, however, we stick to the case ${\cal F}_t = \{X_t,\, X_{t-1}, \ldots\}$. }
For $\{X_t\}$ to be a martingale, all of the relevant information about the future concentrates in the most recently observed value, capturing the notion that a forecast should capture all that we know about $X_{t+1}$.  An immediate consequence of \eqn{evx} is that the differences, or changes, of a martingale resemble independent random variables. In particular, martingale differences $W_{t} = X_t - X_{t-1}$ have mean zero and are uncorrelated: 
\ble
     \ev \, W_t = 0 \quad \mbox{  and  } \quad \Cov(W_s, \, W_t) = 0  \mbox{ for } s \ne t\;.
\qle{mp}
The most well-known example of a martingale is an unbiased random walk.  A random walk accumulates a sum of independent random variables, as in the standard discrete Brownian motion:
\be
         B_t = \sum_{j=0}^t \ep_j = B_{t-1} + \ep_t, \qquad \ep_i \iid N(0,\, 1)  \;.
\qe
The differences $B_t - B_{t-1} = \ep_t$ are, by construction, independent and normally distributed.

Most time series are not martingales.  For example, stationary ARMA time series models are not martingales.  For instance, the first-order autoregression, or AR(1) model, in which
\ble
  Y_t = \rho\, Y_{t-1} + \ep_t \; ,  \quad |\rho| < 1, \; \ep_t \iid N(0, \sigma^2_\ep) \;,
\qle{ar1}
is not a martingale (unless $\rho = 1$, in which case it is a random walk).  For stationary models, the prediction of $Y_t$ given previous values shrinks $Y_{t-1}$ toward the mean of the process at zero, 
\ble
  \yhat_{t|t-1} = \ev \, (Y_t \given Y_{t-1},\, Y_{t-2},\,  \ldots) = \rho \, Y_{t-1} \;.
\qle{yhat-ar1} 
Stationary models such as \eqn{ar1} are mean-reverting whereas a random walk wanders freely.

Although a stationary time series is not a martingale, it is easy to define a martingale from its forecasts.  There's a hint of a martingale in the one-step-ahead prediction $\yhat_{t|t-1}$.  This prediction is a conditional expectation of $Y_t$ given the past, just like that which appears in the definition of a martingale \eqn{evx}.  The martingale structure becomes obvious through a sequence of back-substitutions.  Starting from \eqn{ar1}, plug in the expression from the previous point in time:
\begin{eqnarray}
  Y_t &=& \ep_t +  \rho\, Y_{t-1}           \cr
        &=& \ep_t +  \rho\, \left( \ep_{t-1} + \rho\,Y_{t-2} \right) \cr 
        &=& \ep_t +  \rho\, \ep_{t-1} + \rho^2 Y_{t-2}  \cr 
        &=& \ep_t +  \rho\, \ep_{t-1} + \rho^2 \left( \ep_{t-2} + \rho\,Y_{t-3} \right) \cr 
        &=& \ldots \cr
        &=& \sum_{j=0}^\infty \rho^j \, \ep_{t-j}   \;.
\label{eq:geom}
\end{eqnarray} 
That is, a first-order autoregression can be represented as a geometric average of prior errors. The last expression in \eqn{geom} holds if we think of our data as part of an infinitely long time series; otherwise for a series that begins with $Y_1$, the sum terminates with $\rho^{t-1} Y_1$.

A martingale emerges when we consider how the optimal prediction of $Y_T$ changes as the data approach the target time $T$.  Keep in mind that the martingale describes a sequence of forecasts of a single future value $Y_T$.  The target being forecast does not move (as when extrapolating forecasts farther out in time); instead it is the data available to the forecaster that changes.  Because $\ep_t$ is independent of $Y_s$ for $s < t$, the intermediate expressions building up to \eqn{geom} imply that, for $t \le T$,
\begin{eqnarray*}
   \yhat_{T|t} 
     &=& \ev\left( Y_T \given Y_s, s \le t \right)      \cr
     &=& \ev\left( \ep_T + \rho \, \ep_{T-1} + ... + \rho^{T-t-1} \ep_{T-t-1}  + \rho^{T-t} Y_t 
             \given Y_s, s \le t \right)      \cr
    &=& \rho^{T-t} Y_{t}
\end{eqnarray*}
The resulting sequence of forecasts $\yhat_{T|1},\, \yhat_{T|2},\, \ldots\;, \yhat_{T|T-1},\; \yhat_{T|T} = Y_T$ that ``evolve'' as $t$ approaches $T$ is a martingale,
\begin{eqnarray}
   \ev\left( \yhat_{T|t} \given \yhat_{T|t-1}, \yhat_{T|t-2}, \ldots \right) 
   &=& \ev\left( \rho^{T-t} Y_{t} \given \rho^{T-t+1} Y_{t-1}, \rho^{T-t+2} Y_{t-2}, \ldots \right) \cr
   &=& \rho^{T-t} \ev\left( Y_{t} \given Y_{t-1}, Y_{t-2}, \ldots \right) \cr
   &=& \rho^{T-t} \left(\rho Y_{t-1}\right)  \cr
   &=& \yhat_{T|t-1} \;.
\label{eq:evolve}
\end{eqnarray}

Because $\yhat_{T|t}$ is a martingale in $t$, the changes in the forecasts as $t \,\rightarrow\, T$ are martingale differences and thus uncorrelated.  Define $ \yhat_{T|0} = \ev Y_t = \mu$, the marginal mean of the process (assuming stationarity), and note that $\yhat_{T|T}=Y_T$.  We can then decompose $Y_T$ as a sum of uncorrelated random variables obtained from a telescoping sum of the martingale differences,
\ble
   Y_T = \yhat_{T|T} = \sum_{t=1}^T (\yhat_{T|t} - \yhat_{T|t-1}) \;.
\qle{tele}
This representation then allows us to write the variance of $Y_T$ as
\begin{eqnarray}
   \Var(Y_T) &=& \Var\left( \yhat_{T|T} \right) \cr
                   &=& \Var\left( \sum_{s=1}^T (\yhat_{T|s} - \yhat_{T|s-1})\right)  \cr
                   &=& \sum_{s=1}^T \ev(\yhat_{T|s} - \yhat_{T|s-1})^2 \cr
                   &=& \sum_{s=1}^T \sigma_s^2  \;.
\label{eq:televar}
\end{eqnarray}

We can visualize this algebra to make the process more intuitive.  As illustrated in equation \eqn{geom}, a stationary ARMA model can be expressed as a weighted sum of prior error terms, $Y_T = \sum_j w_j \ep_{t-j}$ where $\sum_j w_j^2 < \infty$.  Consider, for example, the forecast $\yhat_{T|T-3}$.  The forecast consists of summands that are known at time $T-3$, and the rest determine the error of that forecast:
\be
   Y_T = \ep_T + w_1 \, \ep_{T-1} + w_2\, \ep_{T-2} +
                \underbrace{w_3 \,\ep_{T-3} + w_4\,\ep_{T-4} + w_5\,\ep_{T-5} + w_6\,\ep_{T-6} + \cdots}_
                {\yhat_{T|T-3}} \;.
\qe
At time $T-2$, we learn the next component of the sum,
\be
   Y_T = \ep_T + w_1\,\ep_{T-1} +  \underbrace{w_2\,\ep_{T-2} +
               w_3\,\ep_{T-3} + w_4\,\ep_{T-4} + w_5\,\ep_{T-5} + w_6\,\ep_{T-6} + \cdots}_
                {\yhat_{T|T-2}} \;.
\qe
Hence $ \yhat_{T|T-2} = w_2\,\ep_{T-2} + \yhat_{T|T-3}$. Since the $\ep_t$ are independent of each other, it follows that $\ev( \yhat_{T|T-2} \given {\cal F}_{T-3}) = \yhat_{T|T-3}$.  The variance components in \eqn{televar} are easily identifiable as $\sigma_s^2 = \sigma_\ep^2 \, w_s^2$. 
Researchers such as \citet{heath94} and subsequent authors \citep[such as][]{toktay01} use this approach to simulate an arbitrary demand forecasting system.

\section{ Threshold Martingales }
\label{sec:mart}

Most processes are not martingales -- you have to make them.
The approach used to build the martingale $\yhat_{T|t}$ for an autoregression is a special case of a more general construction. Consider an arbitrary random variable $Z$ and an increasing sequence of sigma fields ${\cal F}_0 \, \subset \, {\cal F}_1 \, \subset \cdots.$  It follows from properties of conditional expectation that the sequence of conditional expectations
\ble
    X_t = \ev( Z \given {\cal F}_t ), \quad t = 0,\,1,\, \ldots
\qle{condMart}
is a martingale.\footnote{Some technical conditions, such as bounded expectation $\ev |X| < \infty$ and measurability of $Z$, are required.  These hold in our application.} This construction is the first example of a martingale given by \citet[][Chap 7]{doob53}. Conditional expectations act much like projections in linear algebra, providing an orthogonal decomposition of a vector into subspaces. Hence, the conditional expected value of $Y_T$ given past observations defines a martingale: $\yhat_{T|s} = \ev\,(Y_T \given Y_s,\, s \le T)$ is a martingale.

To define a threshold martingale, the random variable $Z$ in \eqn{condMart} indicates whether the future random variable $Y_T$ lies below a threshold $\tau$. Rather than take the expected value of $Y_T$ itself, we consider the chance that $Y_T$ lies below  $\tau$.  For modeling a continuous random variable, $\tau$ is a quantile of the initial forecast distribution.  Given past observations up to time $t \le T$, the probability that $Y_T \le \tau$ is 
\ble
      p_t = \pr(Y_T \le \tau \given \{Y_s, s \le t\}) = \ev( \one_{Y_T \le \tau} \given {\cal F}_t)
\qle{pt}
where $\one_{W_T \le \tau}$ is an indicator, a Boolean 0/1 r.v. determined by the success or failure of the associated condition,
\be
      \one_{Y_T \le \tau}= \left\{ \begin{array}{cl}
          1 & \mbox{if }  Y_T \le \tau,     \cr
          0 & \mbox{otherwise.}  
          \end{array} \right.
\qe
${\cal F}_t = \{\ldots, \, Y_{t-1},\, Y_{t}\}$ is the sequence of increasing sigma fields defined by prior $Y_t$.  Notice that $p_T \in \{0,\,1\}$.  Since the elements of  $\{p_t\}$ are conditional expectations of a bounded random variable with respect to increasing sigma fields, $\{p_t\}$ is a martingale.  We denote the mean value of this martingale $\pi = \ev(p_t)$.  If $\tau$ is chosen to be a specific quantile of the initial forecast distribution, then $\pi$ is known.

Because $\{p_t\}$ is a martingale, observed sequences should be calibrated with known mean and total volatility.  By calibrated, we mean that $\ev \, (p_t \given p_s, s < t) = p_s$.  Suppose that we observe multiple realizations of $\{p_t\}$, say $\{p_{t,j}\}$ where $j = 1, \ldots, n$.  Then a scatterplot of $p_{t,j}$ on $p_{s,j}$ ($s < t$) should cluster along the diagonal $x=y$.  In addition, sequence plots of $p_{t,j}$ on $t = 0, \ldots, T$ should (on average over $j$) hover around $\pi$, though -- being a martingale -- there is no mean reversion.  Should the multiple realizations be independent (or uncorrelated), then one can easily compare the sample mean $\ol{p}_t = \sum_j p_{t,j}/n$ to the specified probability $\pi$.  Failing calibration can be viewed as an embarrassing mistake since there are ways of guaranteeing calibration even if a simple probabilistic model did not generate your data \citep{fostervohra96}.   When you can assume IID data, often something as simple as the pool adjacent violator algorithm will generate a calibrated forecast \citep{fosterstine04}.

The quadratic variation of $\{p_t\}$ provides a second diagnostic, primarily because the initial quantile determines the expected total quadratic variation.  The event $\{Y_T \le \tau\}$ defines a Boolean random variable with mean $\pi$ and variance $\pi\,(1-\pi)$.  Differencing $\{p_t\}$ decomposes this variance into contributions from period-to-period changes. The difference of $p_T$ from its mean $\pi$ telescopes ($p_0 = \pi$),
\ble
    p_T  - \pi = \sum_{t=1}^T (p_t - p_{t-1})  \;.
\qle{telescope}
Because $p_t - p_{t-1}$ are martingale differences, they are uncorrelated \eqn{mp}.  Consequently,
\begin{eqnarray}
   \ev(p_T  - \pi)^2
       &=& \ev \left( \sum_{t=0}^T \ev(p_t - p_{t-1} \right)^2  \cr
       &=& \sum_{t=0}^T \ev(p_t - p_{t-1} )^2  \cr
       &=& \pi\,(1-\pi) \;.
\label{eq:ss}
\end{eqnarray}
Should the observed total quadratic variation exceed $\pi(1-\pi)$, the process has excess volatility: the forecast distribution is changing more than it should as the data approach the FTD.  One might also observe smaller than expected variation, though that has been less common in our experience.

In addition to having a known total, the quadratic variation of $\{p_t\}$ reflects how fast information about $Y_T$ accumulates in the forecasts. Define the partial sum
\ble
   S_t = \sum_{s=0}^{t} (p_s - p_{s-1})^2 \;.
\qle{st}
$S_t$ is not a martingale, but we find it useful nonetheless as a description of the dynamics of the underlying information about $Y_T$.  Plots of $S_t$ versus $t=1,\ldots,T$ show the flow of information as more and more becomes known about $Y_T$. Small changes $p_{t}-p_{t-1}$ indicate that little new information has arrived.  A simple adjustment converts the sums-of-squares process $S_t$ into a martingale.  In particular, consider
\ble
  V_t = S_t + p_t\,(1-p_t) \;.
\qle{vt}
To see directly that $V_t$ is a martingale, observe that
\begin{eqnarray*}
  \ev(V_{t+1} \given {\cal F}_t) 
     &=& \ev(S_{t+1} + p_{t+1}(1-p_{t+1}) 
\given {\cal F}_t)                                            \cr
     &=& \ev\left(S_{t} + (p_{t+1} - p_{t})^2 + p_{t+1}(1-p_{t+1}) \given {\cal F}_t\right) \cr
     &=& \ev\left(S_{t} + (p_{t+1}^2-2p_{t+1}p_t+ p_{t}^2) + p_{t+1}-p_{t+1}^2 \given {\cal F}_t\right) \cr
     &=& \ev\left(S_{t} - 2p_{t+1}p_t+ p_{t}^2 + p_{t+1} \given {\cal F}_t\right) \cr
     &=& S_{t} + p_t (1 - p_{t}) \cr   
     &=& V_t
\end{eqnarray*}
Martingale differences associated with $V_t$ have some special structure that is worth pointing out.  The differences  can be written as
\begin{eqnarray}
  V_t - V_{t-1} 
    &=& p_t(1-p_t) - p_{t-1}(1-p_{t-1}) + (p_t - p_{t-1})^2 \cr
    &=& p_t - p_{t-1} - 2 p_{t-1}(p_t - p_{t-1})                  \cr
    &=& (p_t - p_{t-1}) (1 - 2 p_{t-1})                  \cr
    &=& 2 (p_t - p_{t-1}) (\half - p_{t-1})      
\label{eq:dvt}
\end{eqnarray}
Hence, the difference $V_t - V_{t-1}  = 0$ when $p_{t-1} = \half$. The visual effect in a plot of $V_t - V_{t-1} $ on $V_{t-1}$ is striking.

\remark{ 
We have two motivations for the construction of $V_t$.  The first is heuristic.  Split the total sum of squared differences at some point into two sums, $S_T = \sum_{t=0}^k \ev(p_t - p_{t-1})^2$ $+ \sum_{t=k+1}^T \ev(p_t - p_{t-1})^2$.  The first summand is $S_k$, and the second has conditional expectation $p_k(1-p_k)$ given ${\cal F}_k$.  The other motivation relies on familiarity with martingale compensators.  The sum of observed differences of the martingale $p_t$ is often written as $S_t = [p,p]_t$, for which the natural compensator is the expected value $\ip{p}{p}_t$.  We don't know $\ip{p}{p}_t$, but this is also the compensator of $p_t^2$.  Hence by adding $p_t(1-p_t) = p_t - p_t^2$ we subtract the unknown compensator and end up with a martingale,
\be
V_t = S_{t} + p_t (1 - p_{t}) = p_t + ([p,p]_t - \ip{p}{p}_t) - (p_t^2 - \ip{p}{p}_t)
\qe
}

\section{Examples for Autoregressions}

This section illustrates the threshold martingale $\{p_t\}$ within the context of autoregressive models.  We start by simulating $\{p_t\}$ for a correctly specified autoregression, and then illustrate a mixture of autoregressions. In these examples, the $\{p_t\}$ are martingales by construction.  We emphasize
\begin{description}
  \item [Calibration plots] A scatterplot of $p_t$ on any lag $p_{t-j}$ should concentrate along the diagonal; 
  \item[Calibration regression] Coefficients in the regression of 
           $p_t - p_{t-s}$ on $p_{t-s_1}, \ldots, p_{t-s_k}$, for $0 < s < s_1 < \cdots < s_k $ should be zero.
  \item [Cumulative sums of squares] The total variation should average $\pi(1-\pi)$.
\end{description}

\subsection{Correctly Specified Models: One realization}  %

\begin{figure}
\caption{{\it Snapshot of a first-order autoregression \eqn{ar1} with the associated conditional probabilities ${p}_t = P(Y_{40} \le 1 \given {\cal F}_t)$. } The coefficient $\rho = 0.8$ with innovation variance $\sigma^2 = 1$.}
\label{fi:autoregr}
   \centerline{ \includegraphics[width=4.5in, height=4in]{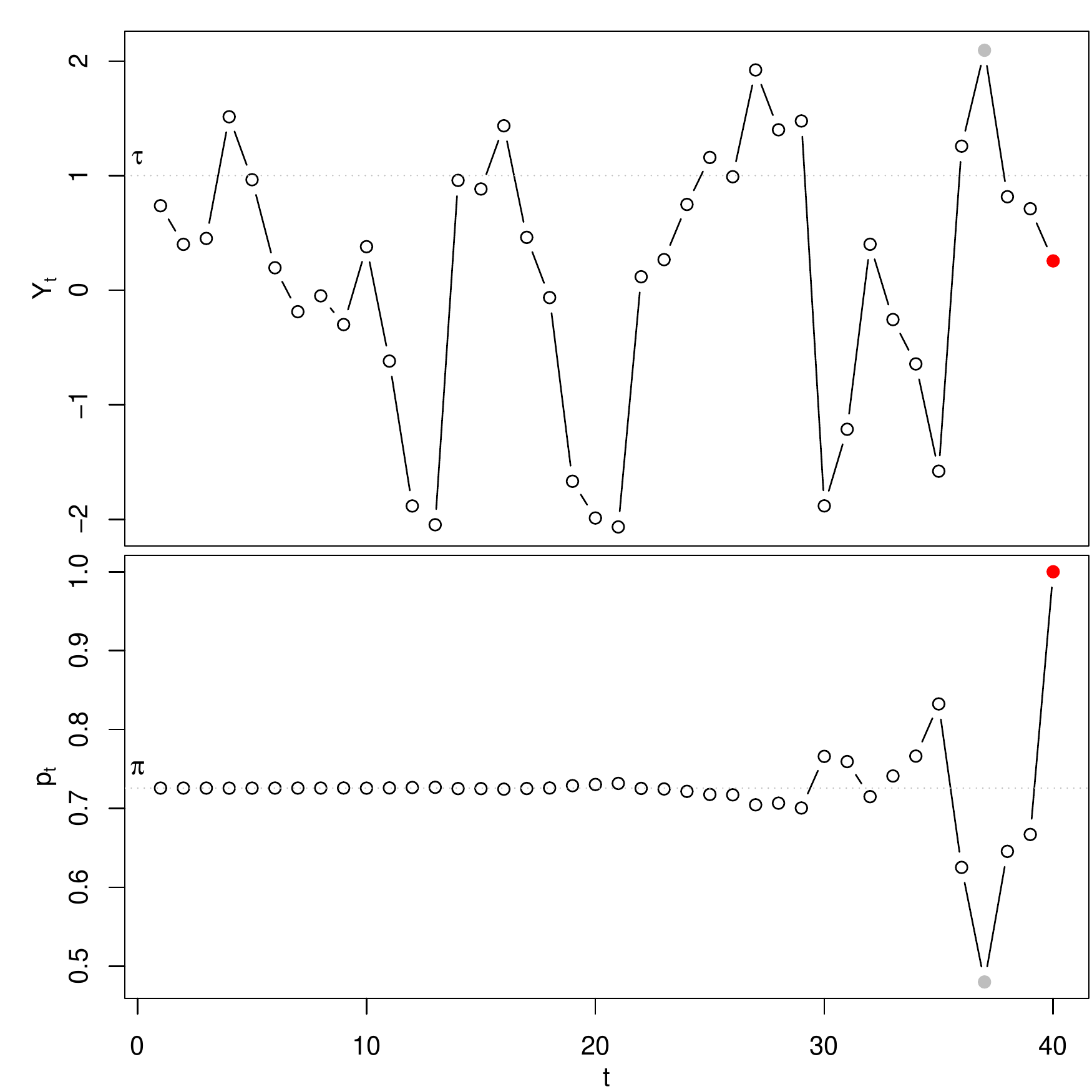} }
\end{figure}

We begin with an illustration of the threshold martingale for a single time series.  Figure \ref{fi:autoregr} graphs $\{p_t\}$ for a segment of a simulated first-order autoregression with coefficient $\rho = 0.8$ and $\sigma_\ep = 1$. The top frame of the figure shows $Y_t$ for $t = 1,\ldots, 40$.  The dotted gray line denotes the threshold $\tau = 1$; the target period $T=40$ lies at the right side of the graph.  The lower frame of Figure \ref{fi:autoregr} shows the threshold probabilities $p_1, \, p_2,\, \ldots, p_{40}$ for the data in the top frame.  The $p_t$ vary around the threshold probability 
\be
   \pi = \pr(Y_t \le 1) = \pr\left( N(0,1) \le \sqrt{1-\rho^2} \right) \approx 0.73 \;.
\qe
The sequence of probabilities is  basically constant for $t  \ll T$ because little is known about where the process will be at $T=40$ other than what is implied by the known stationary distribution, $Y_t \sim N(0, \smfrac{1}{1-\rho^2})$.  At $t=1$, for example, the forecast of $Y_{40}$ is $\rho^{39} Y_1 \approx 0$.  

Variability in the probability martingale reveals how information flows in the underlying process.  Periods of higher variation of the conditional probabilities show that the associated data contain more information about the outcome $Y_T$.    Although the time series $Y_t$ meanders, with several excursions above the threshold of interest ($\tau < Y_T$), $p_t$ remains near constant until around $t \approx 25$, at which point information in the data begins to reveal the most likely position for $Y_T$.   As shown in \eqn{geom}, an autoregression is a weighted average of prior random inputs, and in this case $Y_T = \sum_t 0.8^j \ep_{T-j}$.  Observing $Y_t$ reveals the contribution $0.8^{T-t} \ep_t$ which is quite small until $t$ approaches the target time.  The smallest probability $p_t$ occurs at $t=37$ (gray points in the figure) when $Y_t$ is quite large and $t$ is close to $T$.  The conditional probability that $Y_t \le 1$ increases from $t = 37$ because the subsequent observations decrease.  At the end of this sequence, the target $Y_T \le \tau$ and consequently $p_T = 1$ (red point in the figure).

\begin{figure}
\caption{{\it Probability martingales for several thresholds of the simulated first-order autoregression shown in Figure 1.} The thresholds are $\tau = -0.5,\, 0, \,0.5,\, 1,\, 2$.  Dashed horizontal lines indicate the expected value for each martingale. }
\label{fi:arflow}
   \centerline{ \includegraphics[width=5in, height=3in]{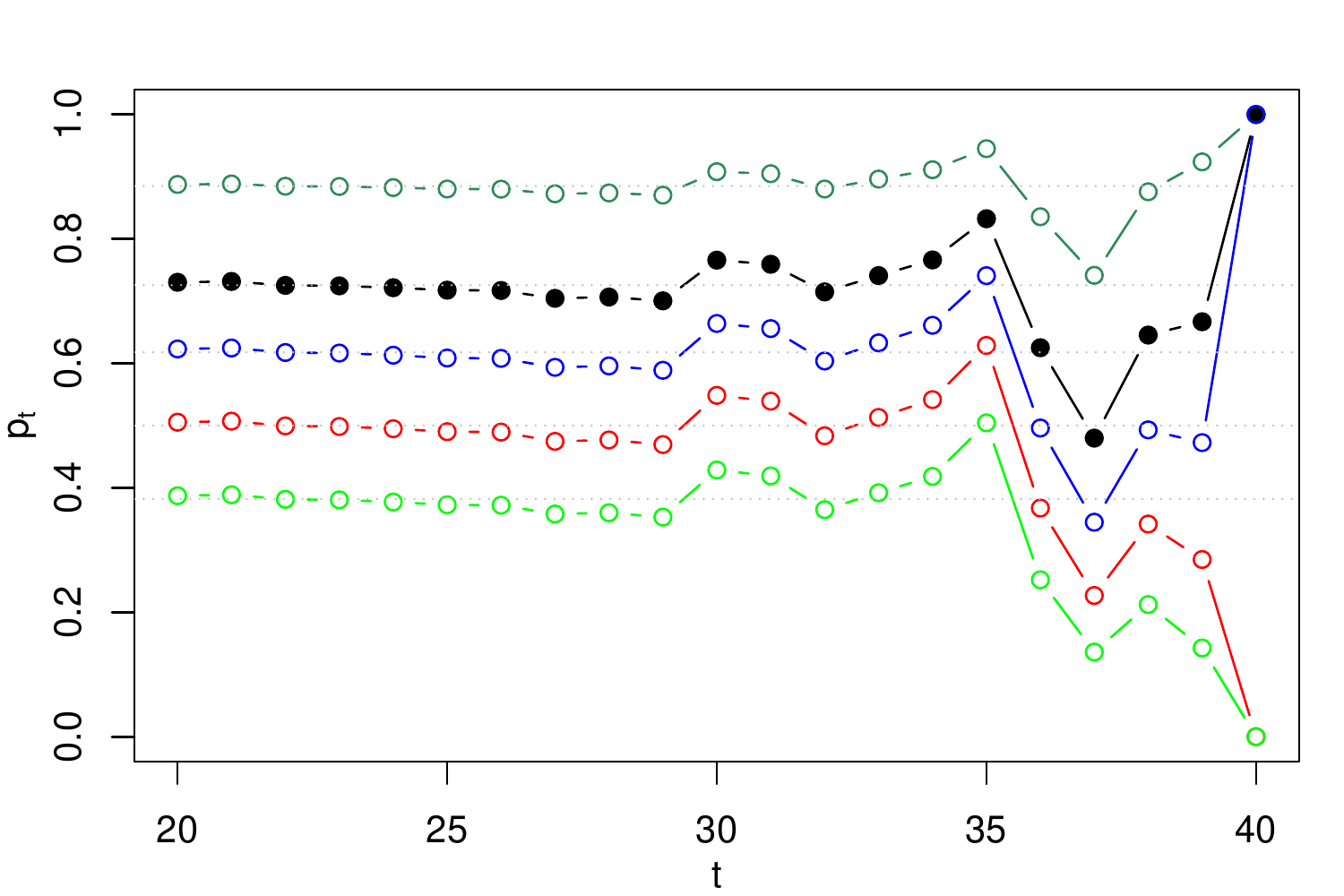} }
\end{figure}

Rather than isolate a single threshold, one can track probabilities associated with several thresholds at once.  The resulting martingale is now a vector of highly correlated probabilities.  Figure \ref{fi:arflow} shows the conditional probabilities for 5 thresholds located at $\tau = -0.5,\ 0,\  0.5,\, 1,$ and 2.  The plot shows larger values $ t \ge 20$ with noticeable variation; prior values are nearly constant as in Figure \ref{fi:autoregr}. The filled points in the figure repeat the conditional probabilities for $\tau=1$ found in Figure \ref{fi:autoregr}.  Notice that the conditional probabilities diverge as $t$ nears the target $T$.  The final value $Y_{40} \approx 0.26$ exceeds two of the thresholds.  The conditional probabilities for $\tau = -0.5$ and $\tau = 0$ both end at 0 whereas those for the larger thresholds end at 1.

\begin{figure}
\caption{{\it Accumulation of sums of squares of $\hat{p}_t - \hat{p}_{t-1}$ for (a) one realization of a first-order autoregression \eqn{ar1} and (b) infinitely many realizations.} Results in black set $\rho=0.8$ and those in gray have $\rho = 0.995$.}
\label{fi:arss}
   \centerline{ \includegraphics[width=2.5in]{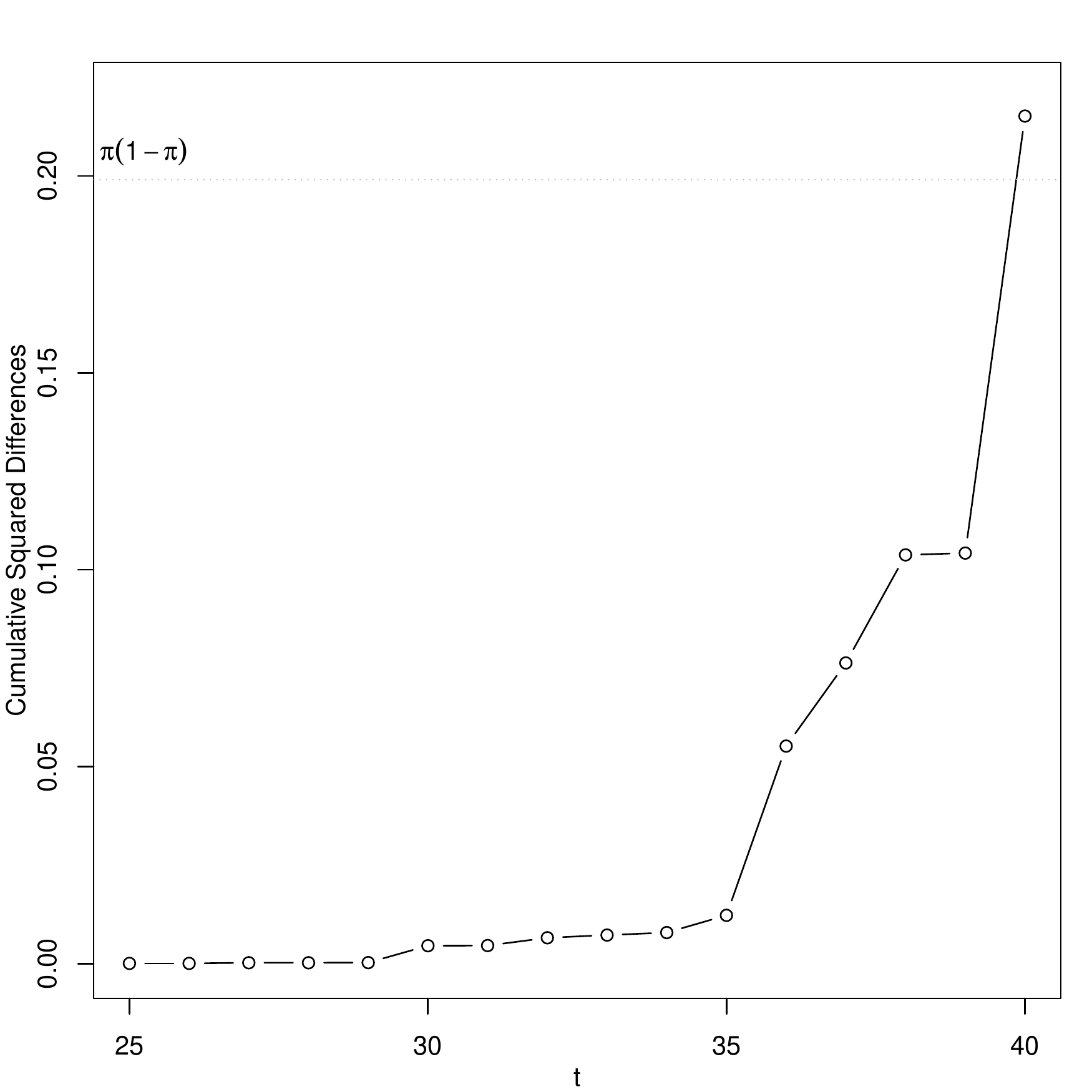} 
                      \includegraphics[width=2.5in]{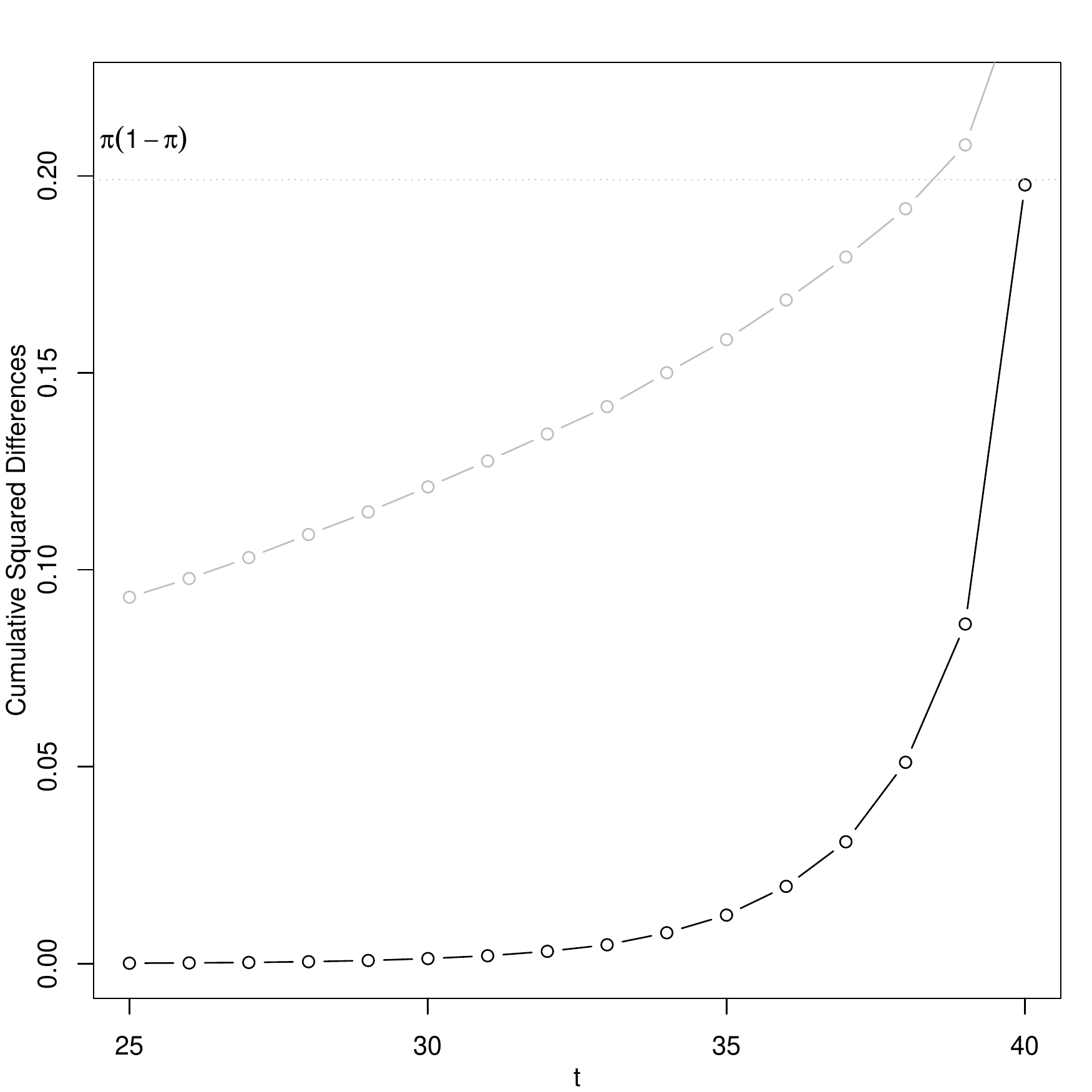}
   }
\end{figure}

The accumulating variability of $\{p_t\}$ reflects how fast information about $Y_T$ accumulates.  The rate at which variation accumulates as measured by $S_t$ defined in \eqn{st} depends on the memory of the underlying process.   To illustrate, the gray curve in Figure \ref{fi:arss}(b) shows $S_t$ for a process with $\rho = 0.995$; realizations with $\rho \approx 1$ resemble a random walk and have a much larger marginal variance with longer excursions from the mean at 0.  The resulting growth in the cumulative sum of squares is nearly linear; observations in the distant past influence the prediction of $Y_T$ almost as much as those closer in time. The level of the gray curve is higher and its total is larger than that for the process with $\rho = 0.8$.  If $\rho = 0.995$, then $\pr(Y_T \le 1) \approx 0.54$.

\begin{figure}
\caption{{\sl The conditional probabilities are calibrated for both autoregressive processes.} Results in black are from realizations with $\rho=0.8$ and those in gray with $\rho = 0.995$. The black line in the figure in the diagonal of the plot; the nearly indistinguishable red line in the figure is the least squares line fit to the gray points.}
\label{fi:ar_calib}
   \centerline{ \includegraphics[width=3in]{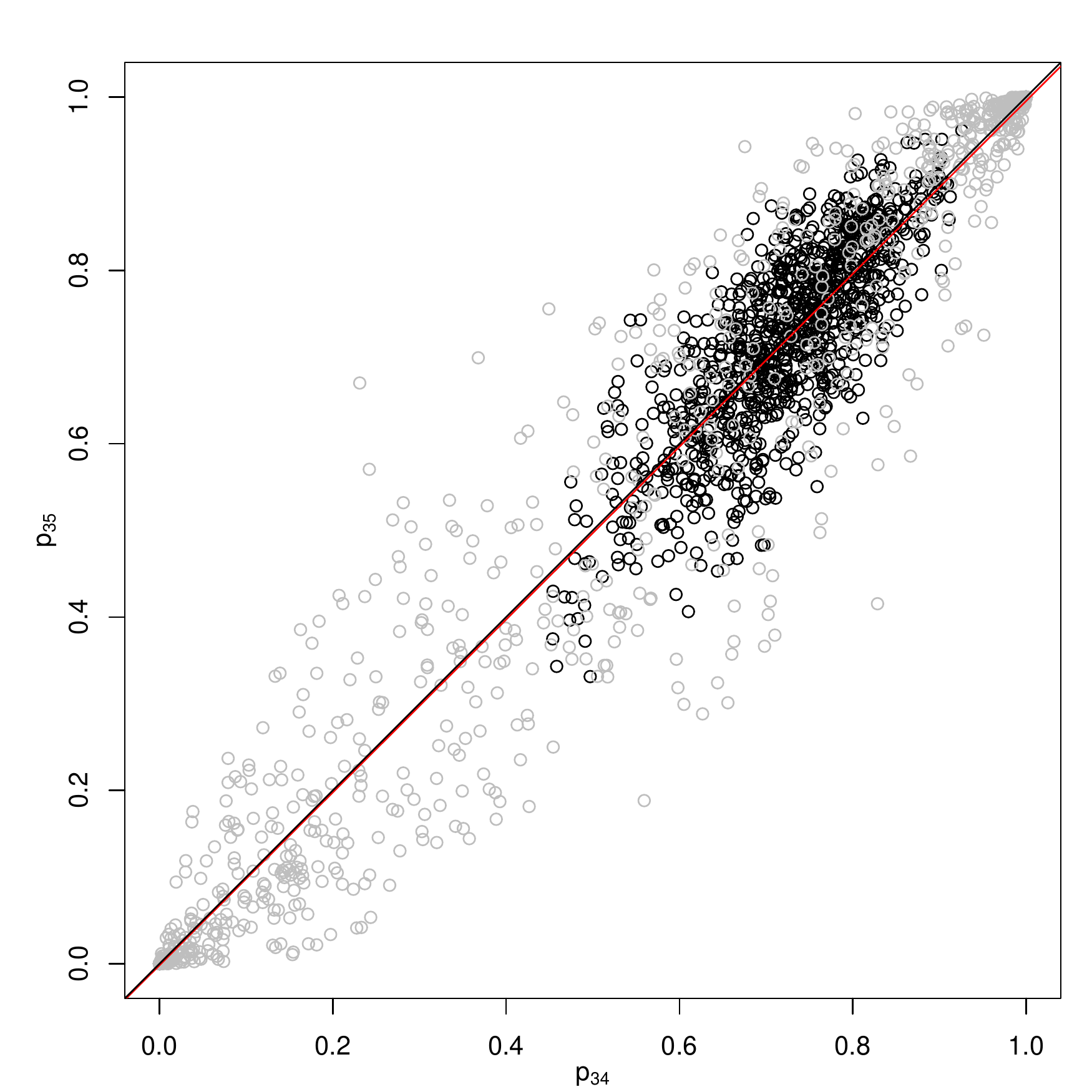} 
   }
\end{figure}

Verifying the calibration $\{p_t\}$ does not depend on the choice of $\tau$ or $\rho$. Because $\{p_t\}$ is a martingale, the average value of the process at the next point in time, $p_{t+1}$, is the most recent value, $
  \ev( p_{t+1} \given {\cal F}_{t} ) = p_t $ {\em regardless}  whether we know $\rho$ or $\pi$.  Hence, if the threshold probabilities form a martingale, a scatterplot of $p_{t}$ on $p_s$, $s < t$ should concentrate on the diagonal line $y = x$.  Figure \ref{fi:ar_calib} illustrates this calibration. This scatterplot graphs $p_{35}$ on $p_{34}$ derived from 400 independent replicates of the autoregressions with either $\rho = 0.8$ (black) or $\rho = 0.995$ (gray).  Although from different processes, both sets of coordinates align on the diagonal.  We can clearly distinguish $p_t$ generated by the different processes, but both are calibrated.  The figure includes the least squares line fit to the points with $\rho=0.995$; the fitted line is nearly indistinguishable from the diagonal.

A family of multiple regression models nicely summarizes the martingale structure. Consider the regression of $p_t$ on $k$ preceding probabilities $p_{t-1}, p_{t-2}, \ldots, p_{t-k}$.  The only nonzero term is the prior probability,
\be
   \ev (p_{t} \given p_{t-1}, \ldots,\, p_{t-k}) = p_{t-1} \;.
\qe
Only the coefficient of the first lagged probability differs from zero, and it has coefficient 1.  More generally, in the regression for which the most recent lag is $s_1 < s_2 < \cdots < s_k$,
\be
   \ev (p_{t} \given p_{t-s_1}, \ldots,\, p_{t-s_k}) = p_{t-s_1} \;.
\qe
Claims such as these are easily tested by considering the martingale differences for which the conditional mean is zero, as in
\ble
   \ev (p_{t} -  p_{t-s} \given p_{t-s_1}, \ldots,\, p_{t-s_k}) = 0 \;.
\qle{gamma_regr}
A regression of $p_t - p_{t-s}$ on any collection of prior values at lags $s_j > s$ should find no signal.  Indeed, this property extends more generally to {\em any} function of the prior values, not just linear functions.  That is, $\ev (p_{t} -  p_{t-s} \given g(p_{t-s_1}, \ldots,\, p_{t-s_k})) = 0$ for any measurable function $g(\cdot)$.

\subsection{Correctly Specified Models: Ensemble Diagnostics}  %

\begin{figure}
\caption{{\sl Probability martingales for an ensemble of 250 simulated autoregressions with common threshold probability but  varying levels of dependence sampled from the inset distribution.} The threshold for each realization is chosen so that all have the same mean $\pi=0.75$. Colors distinguish sequences ending in 0 (green) or 1 (gray).}
\label{fi:esmProb}
   \centerline{ \includegraphics[width=5in,height=3in]{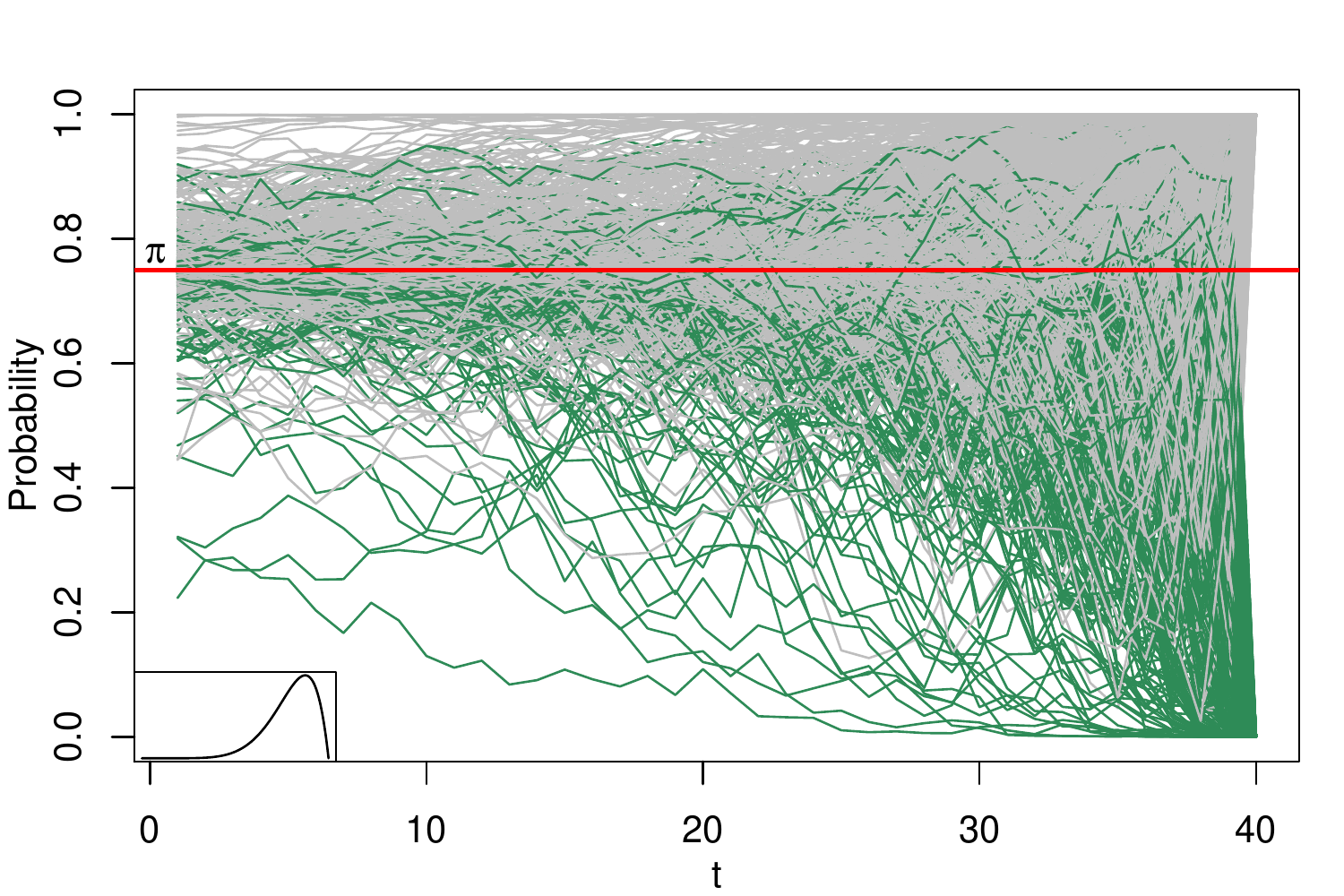} 
   }
\end{figure}

The properties of a martingale vary widely from realization to realization.  For example, sometimes the terminal value is less than the
threshold, $Y_T \le \tau$, and sometimes not. To illustrate the variation from case to case, Figure \ref{fi:esmProb} shows 250 
independent realizations of threshold martingales $p_{t,j}$ defined by $j = 1, \ldots, N=250$ autoregressions with varying coefficients 
but a common target probability $\pi$. Throughout this section, the subscript $j$ identifies that the value is associated with the $j$th realization.  These mimic what one would like to find, for example, when comparing demand forecasts for different products; each product has a different threshold, but the probability $\pi$ is the same for all.  For this example, the simulated coefficients of the AR(1) processes are random draws from a Beta distribution, $\rho_j \iid \mbox{Beta}(8, \, 2)$.  This is a skewed distribution on the interval $[0,\,1]$ with mean 8/10 and SD $\approx$ 0.12. The inset in Figure \ref{fi:esmProb} shows the probability density.  For the $j$th realization, the threshold $\tau_j$ satisfies $\pr(Y_{T,j} \le \tau_j) = 0.75$, which is close to the threshold in the prior examples of a single time series.  By choosing a specific threshold for each process, the expected value of every threshold martingale is  $\pi = \ev\,\one_{Y_{T,j} \le \tau_j} = 0.75$, even though the underlying time series $Y_{t,j}$ have varying levels of dependence.  In the figure, sequences that end at $p_{T,j} = 0$ (realizations for which $\tau_j < Y_{T,j}$, holding for about 25\% of the time series) are colored green, whereas the more numerous series ending at $p_{T,j} = 1$ are colored gray. By and large, small values of $p_{t, j}$ that occur early in the sequence for small $t$ suggest a sequence that will terminate at 0, but quite a bit of variation occurs close to $T$.  The colors of the sequences in the figure are not previsible and would not be revealed until the final value is observed.

\begin{figure}
\caption{{\sl Average cumulative sum-of-squared differences $S_t$ for the collection of autoregressions with varying dependence.} Gray lines show sums-of-squares for individual realizations. Colors distinguish sequences ending in 0 (green) or 1 (gray).}
\label{fi:esmSS}
   \centerline{ \includegraphics[width=5in, height=3in]{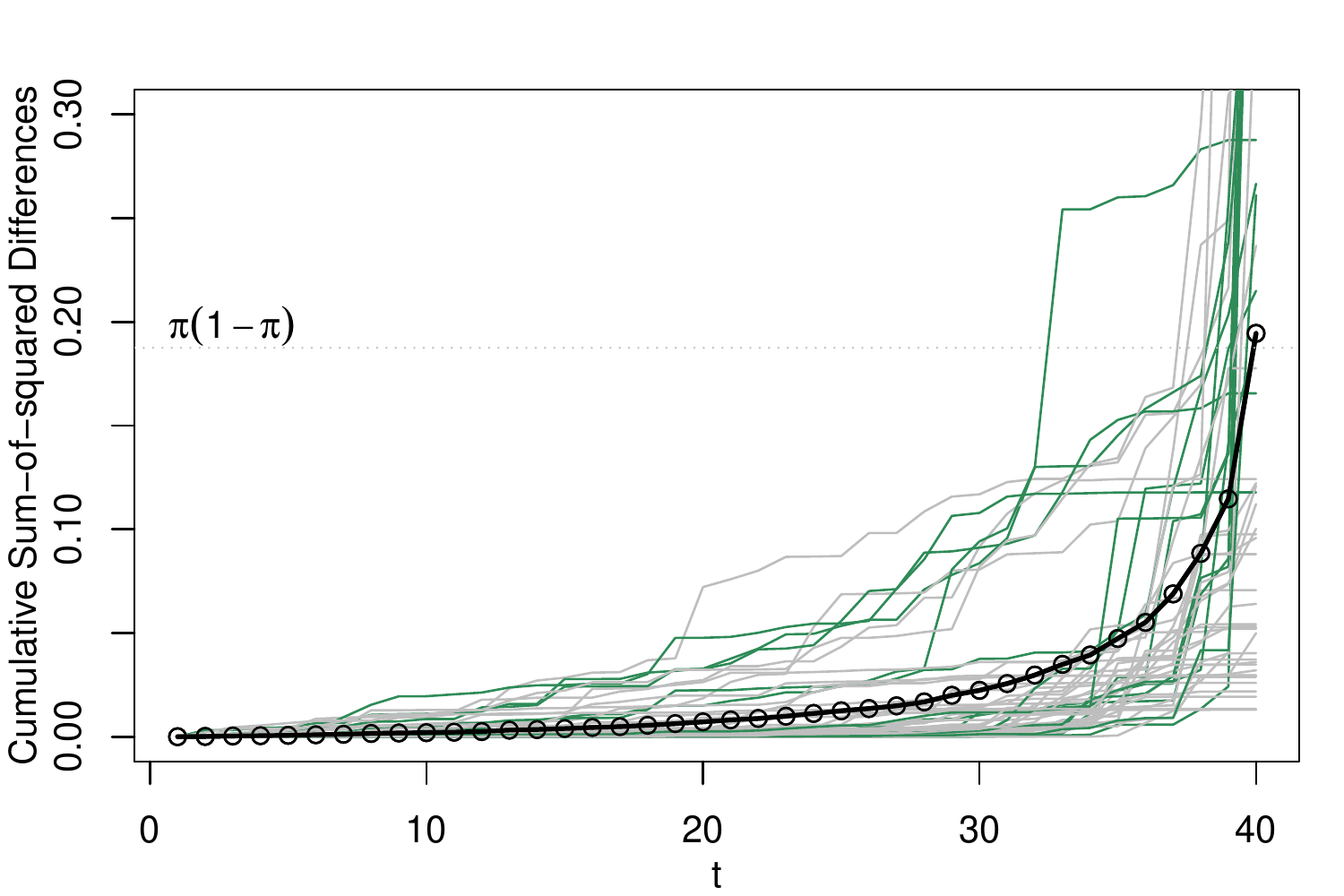} 
   }
\end{figure}

Varying the threshold to obtain a fixed probability $\pi = \pr(Y_{T,j} \le \tau_j)$ allows us to combine the variation across series. The accumulated sum-of-squared martingale differences 
\be
     d_{t,j}^2 = (p_{t_j} - p_{t-1,j})^2
\qe
sum to $\pi(1-\pi)$, on average,
\be
    \ev  \sum_t d_{t,j}^2 = \pi(1-\pi) \qquad \forall j.
\qe
Although the variation accumulates at different rates depending on $\rho_j$ (see Figure \ref{fi:arss}), the total sum-of-squares is the 
same, on average.  Figure \ref{fi:esmSS} shows the average sum-of-squares for the same 250 realizations as shown in Figure 
\ref{fi:esmProb}.  Even after averaging over these realizations, the final total is noticeably less than $\pi(1-\pi)$ due to sampling 
variation.  The light gray lines in the figure show the sums-of-squares for 20 randomly selected series; these remind one of the large 
variation across the series.  The total variation varies from near 0 to far larger than the expected value $\pi(1-\pi)$; some realizations 
are near constant whereas others are far more volatile.

Verifying that the mean of the total accumulated variation matches $\pi(1-\pi)$ in this example using a statistical test is a simple calculation.  We only need the mean and its usual standard error. We have a mixture of autoregressions in this collection of time series, implying in general that squared differences of the associated probability martingales accumulate at different rates, $\ev\,d_{t,j}^2 -  d_{t,k}^2 \ne 0$.  Hence looking at measures of variation at intermediate times $0 < t < T$ would require an adjustment for heteroscedasticity.  The total, however, remains $\pi(1-\pi)$ regardless.  For the 250 time series shown in Figure \ref{fi:esmSS}, the average cumulative value estimates the variance of $p_{T,j} = \one_{Y_T \le \tau_j}$.  The average squared deviation from the target probability is
\be
    \smfrac{1}{250} \sum_{j=1}^{250} (p_{T,j} - 0.75)^2 \approx 0.179,   \qquad SE \approx 0.02, 
\qe
which is well within a standard error of $\pi(1-\pi) = 3/16 = 0.1875$.\footnote{This use of the standard normal-theory test of a sample mean ignores the considerable skewness in the associated binomial distribution. One is advised to take this approach only with a large collection of series.}

\begin{figure}
\caption{{\sl Calibration plots of $p_t$ on $p_s$ $(s < t)$ concentrate along the diagonal, regardless of the separation of $t-s$.}  Each frame shows a diagonal reference line.  None of the departures from the diagonal fit are statistically significant. }
\label{fi:esmCalib}
   \centerline{ \includegraphics[width=3in]{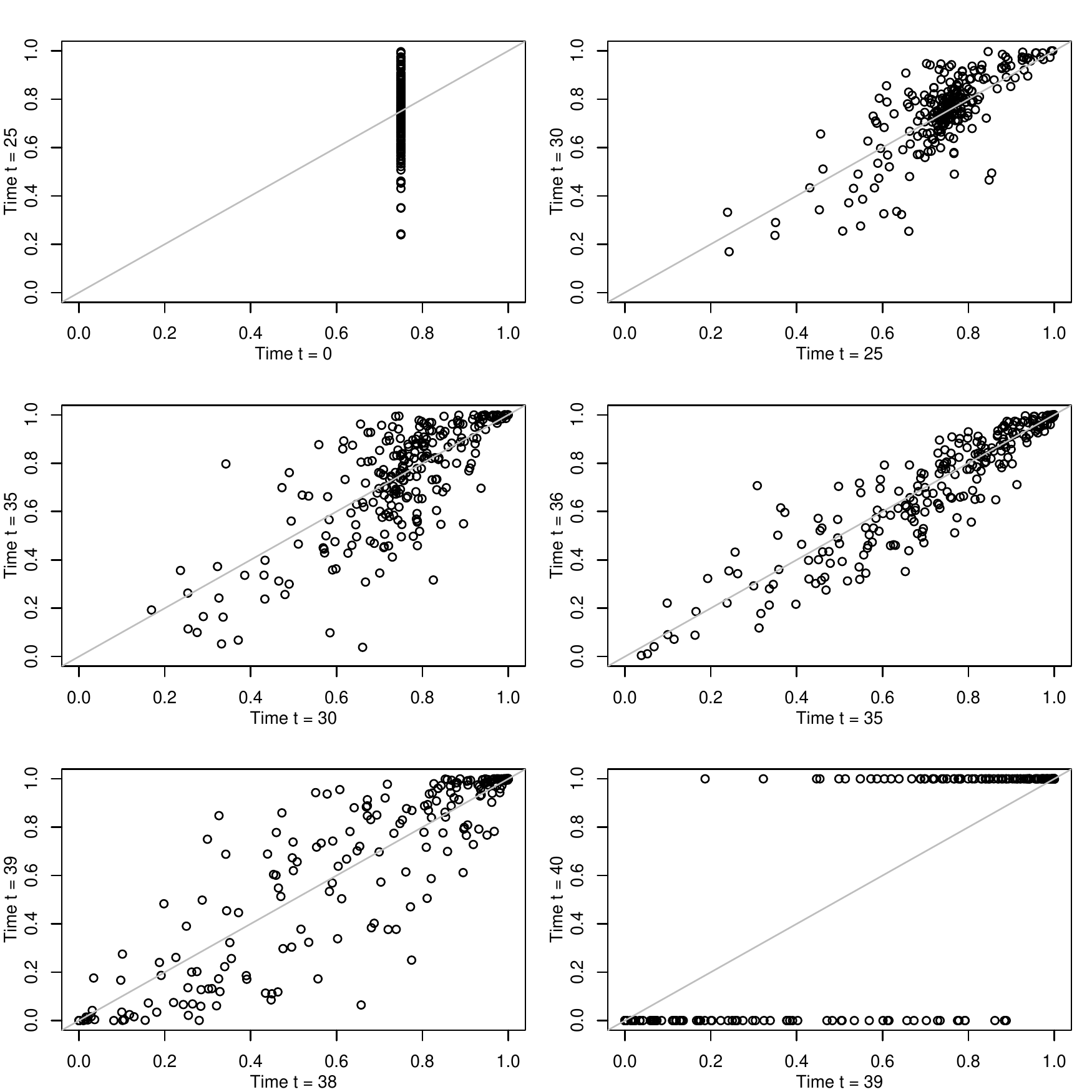} 
   }
\end{figure}

Another useful, visual diagnostic is to inspect the calibration of the probabilities.  As noted in Figure \ref{fi:ar_calib}, the key property of martingales $\ev(X_t \given {\cal F}_s, s \le t) = X_s$ holds regardless of the level of dependence.  Consequently, a scatterplot of $X_t$ on any $X_s,\, s \le t$ concentrates on the diagonal line with slope 1.  The frames in Figure \ref{fi:esmCalib} show several examples, once again based on the probability martingales for the mixture of 250 autoregressions.  The upper frame on the left graphs $p_{25}$ on $p_0 = \pi = 0.75$. All of the series start with $p_0 = 0.75$; the vertical spread shows that the probabilities have begun to spread out by $t=25$.  The remaining plots use larger values of $t$ to highlight regions with more variation.  Reading the plots as lines in a book, the remaining plots graph $p_{30}$ on $p_{25}$, $p_{35}$ on $p_{30}$, $p_{36}$ on $p_{35}$, $p_{39}$ on $p_{38}$, and finally $p_{40}$ (the terminal 0/1 variable) on $p_{39}$.  In each case, the points cluster along the diagonal.  To test the significance of any  deviation from linearity requires adjusting for heteroscedasticity.  Using sandwich estimates of standard error, only the linear term in fifth-degree polynomial fits is statistically significant in these examples.

\begin{figure}
\caption{{\sl Boxplots of the overall reported explained variation for autoregressions fit to the martingale differences.}  Heteroscedasticity distorts the OLS $R^2$ but not the prequential statistic $\widetilde{R}^2$ shown in equation \eqn{adjr2}. }
\label{fi:boxplots}
   \centerline{ \includegraphics[width=3in]{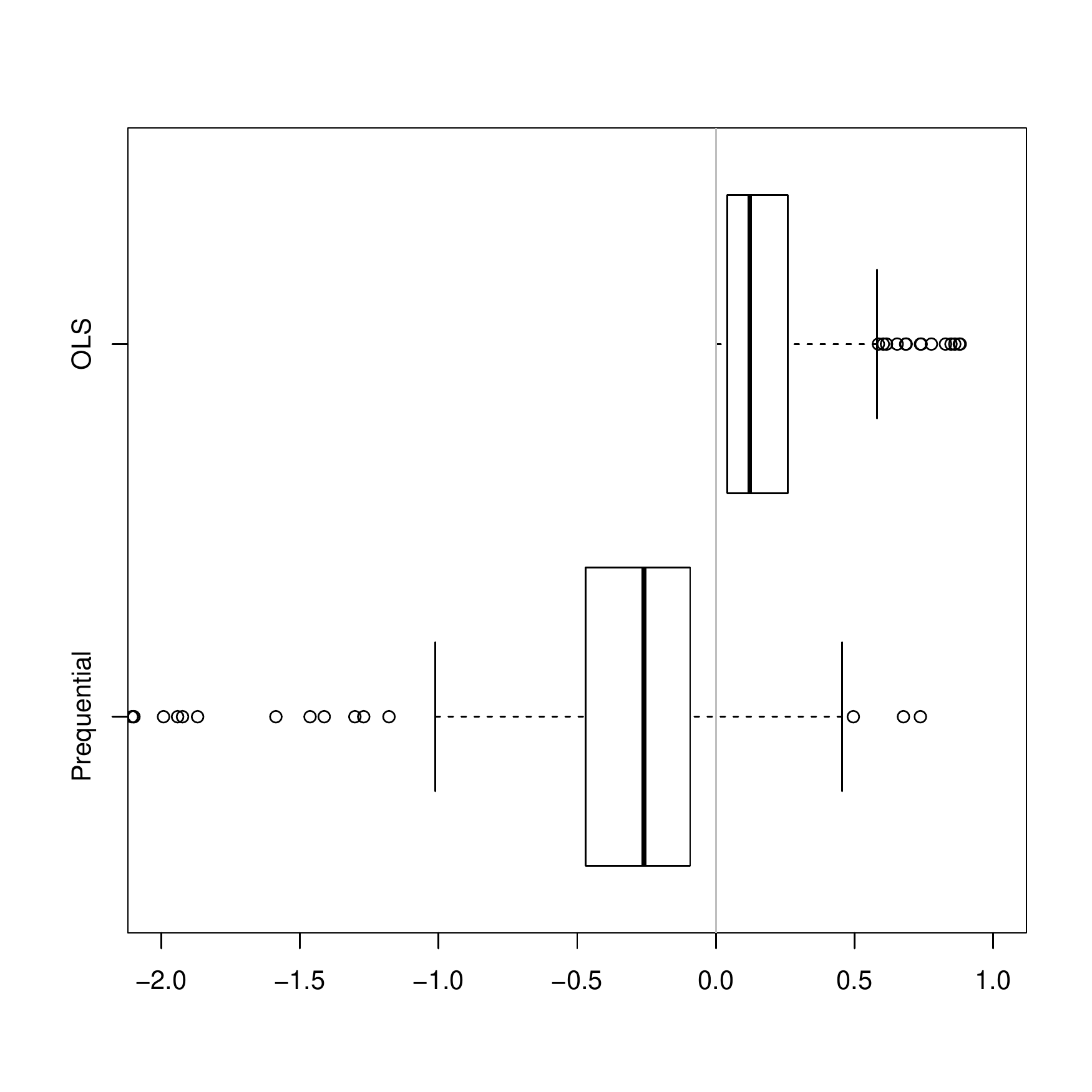} 
   }
\end{figure}

As a final diagnostic, consider the presence of dependence within the differenced probabilities $d_{t,j} =p_{t,j} - p_{t-1,j}$. These are martingale differences, and hence are should be uncorrelated. Calibration measures association between, say, two times, averaging the data {\em across} the replicated series.  Autocorrelation measures dependence between adjacent values {\em within} a series, averaging over time.   
We illustrate a simple diagnostic here. Consider the proportion of variation explained in the regression of $d_{t,j}$ on $d_{t-1,j}, \, d_{t-2,j}, \, \ldots, d_{t-\ell, j}$ for maximum lag $\ell = 4$.  The changing variation of the probabilities fools a na\"ive estimate of this goodness-of-fit.  As seen in Figure \ref{fi:esmSS}, most of the variation in each series occurs near the final time $T$; each $p_t$ is near constant initially with most changes occurring near the end.  The resulting regression of $d_{t,j}$ on lags has a few highly leveraged points, complicating inference. This form of heteroscedasticity inflates the $R^2$ statistic of the complete-data OLS regression. Figure \ref{fi:boxplots} shows a boxplot of these statistics from fitting OLS regressions; these statistics incorrectly indicate that lags explain statistically significant variation, violating the martingale condition.   We avoid these complications by estimating the fit of the regression of $d_{t,j}$ on $d_{t-1,j}, \, d_{t-2,j}, \, \ldots, d_{t-\ell, j}$ using the prequential approach \citep{dawid84, dawid92}.  The reported statistic is
\ble
  \widetilde{R}_j^2 = 1 - \frac{\sum_{t = t_0}^T (d_{t,j} - \hat{d}_{t,j})^2}
                                     {\sum_{t = t_0}^T d_{t,j}^2}\;, j = 1,\ldots,N,
\qle{adjr2}
where the prequential predictor is
\be
  \hat{d}_{t,j} = \hat{b}_{0t} + \hat{b}_{1t} d_{t-1,j} + \cdots + \hat{b}_{\ell,t} d_{t-\ell,j} \;.
\qe
The coefficients in $\hat{d}_{t,j}$ are estimated for each series to minimize the fit up to time $t$,
\be
   \hat{b}_{j,t} = \arg \min_b \sum_{t=\ell+1}^{t-1}  \left(
                  d_{t,j} - (b_{0t} + b_{1t} d_{t-1,j} + \cdots + b_{t-\ell,j} d_{t-\ell,j}) 
                  \right)^2 \;.
\qe
Prequential regression predicts $d_{t,j}$ using a model fit to data {\em prior} to time $t$, recursively updating the fit over time.  Over 90\% of the $\widetilde{R}^2$ are negative, indicating (correctly) that the lagged variables are not predictive. Figure \ref{fi:boxplots} contrasts the distribution of $\widetilde{R}_j^2$ to that given by the usual estimator.  On average, the models are not predictive.  Alternatively, one could avoid the prequential approach and compute t-statistics and an overall F-like statistic using a sandwich style estimator of variances for OLS estimates.

\section{ Application: Basketball Scores }
\label{sec:bb}

This section illustrates the use of martingale diagnostics to evaluate forecasts from two models that predict the winning team in a professional basketball game. The forecast distribution of these models is particularly simple, namely the probability of the home team winning a basketball game. One of these models is intentionally crude whereas the second is more predictive; that said, neither should be taken as a serious competitor to modern ML sports models \citep[\eg][]{shisong19, shisong20}. We also illustrate the use of a martingale filter (described in the Appendix) that reduces forecast volatility to the expected level while simultaneously improving the accuracy of forecasts.

The data for our example span four seasons of play in the NBA. Early quantitative models for the evolution of the probability of a home win made due with much less data.  For example, \citet{stern94} used scores at the end of each quarter of 493 games during the 1992 season to build a Brownian motion model for scoring.  (Each game lasts 48 minutes, divided into 4 quarters of 12 minutes each.) Technology has changed that situation.  The data we study give the score every 15 seconds during 4,622 regular season NBA games from 2015 through 2019.  The data from the 2015-16, 2016-17, and 2017-18 seasons form the training sample, and the 2018-19 season comprises the test sample.  After removing games with errors in the scoring, the three seasons of training data span 3,461 games, and the test season has 1,161 games.  The models make no use of the season or team names and treat all games during the four seasons equivalently.\footnote{We obtained these data  from the Kaggle web site https://www.kaggle.com/schmadam97/. The data provide a tabular play-by-play record of each game.  We limit our analysis to regular season games ending during the standard 48 minutes, omitting overtime and playoff games. We excluded the two most recent seasons that were shortened or influenced by Covid-19; for example, the home-team advantage weakened in the 2019-2020 season in the absence of court-side fans. }

We use the following notation to describe our models.  Randomly label the two teams playing in any game Team A and Team B.  The binary variable $Y_j \in \{0,1\}$ denotes whether Team A wins the $j$th game, $Y_j = 1$ if Team A wins and $Y_j=0$ if Team B wins. The subscript $j = 1, 2, \ldots,4622$ identifies the game throughout.   Without further information, the probability that Team A wins game $j$ is
\be
   p_{-1,j} = \ev(Y_j \given {\cal F}_{-1}) = 0.5  \;, \quad \mbox{ where } \quad {\cal F}_{-1} = \{\varnothing, \Omega\}
\qe
denotes the trivial sigma field.  Now suppose that the game is about to begin and the home team team is identified; the binary random variable $H_j = 1$ if Team A is the home team and 0 otherwise.  This random variable defines the initial sigma field ${\cal F}_{0,j} = H_j$, and denote
\be
   p_{0j} = \ev(Y_j \given {\cal F}_{0}) =  \ev(Y_j \given H_j) \;.
\qe
By symmetry, assume that Team A is the home team. Denote the times when the score is recorded by $t_0=0$, $t_1$=0:15, $t_2$= 0:30, $\ldots$, $t_{191}$=47:45, $t_{192}$=47.59, just before the game ends.  These scores define the (time $\times$ game) matrix 
\be
   X_{ij} = (\mbox{Home Score} - \mbox{Away Score}) \quad  \; i =0, \ldots, 192, j=1, \ldots, 4622,
\qe
$X_{0j} = 0$ for all games; all that is known about the game at the tip-off is the identifier of the home team.  Subsequent rows in $X$ add the difference in the score.  The sigma field ${\cal F}_{ij} = \{H_j,\, X_{s}: s \le t_i\}$ identifies the home team and the home minus away scores up to time $t_i$ in game $j$.  For each game, define the discrete series 
\be
     p_{ij} = \ev(Y_j \given {\cal F}_{t_i,j}).
\qe
By construction, the sequence $\{p_{ij}\}$ is a discrete martingale in $i$.  The illustrative models defined next estimate these probabilities.

The first model uses only the current score difference and the identity of the home team to predict the winner. The model computes the probability of a home win as
\ble
  \hat{p}_{ij}^{(1)} = g(\hat\alpha_0 + \hat\alpha_1 X_{ij}) \;,
\qle{p1}
where $g(x)$ is the logistic function, $ g(x) = \frac{1}{1+e^{-x}}$. We estimate the two parameters $\hat\al_0 =  0.276\, (s.e. \, 0.004)$ and $\hat\al_1 = 0.1681\, (s.e. \, 0.0006) $ in \eqn{p1} from the training data, pooling score differences from all of the games and fitting a single model. These standard errors are too small as they treat every observation, both within and across games, as independent.  
A passing familiarity with basketball suggests that this logistic regression has a serious flaw: it weights all score differences equally regardless of the time left in the game.  The probability of a home win if the score is tied remains $1/(1+\exp(-\hat\al_0)) \approx 0.568$ regardless of the time remaining.  Or, a home lead of 5 points in the first quarter is just as predictive of a home win as a lead of 5 points with 10 seconds left in the game.  As a result, the probability of winning based on early score differences is more volatile than it should be.

\begin{figure}
\caption{\it The polynomial $\hat{w}_7(t)$ varies the coefficient of the score difference over the course of a game. Early score differences have small weight, whereas those near the end of a game are more important. Points marked $\times$ in the figure show slopes of cross-sectional models fit at distinct points in time.}
\label{fi:w_of_t}
   \centerline{ \includegraphics[width=4.5in, height=3in]{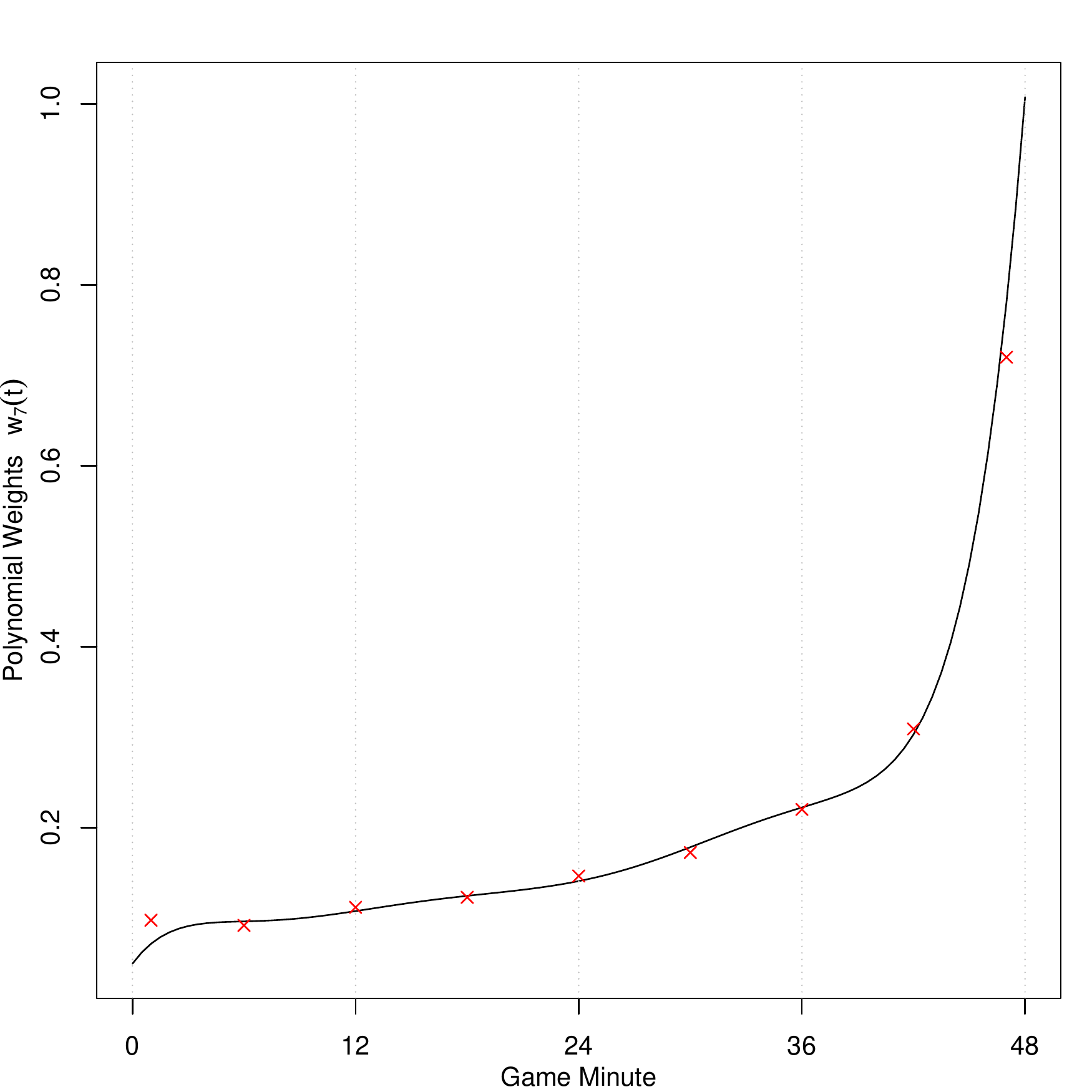} }
\end{figure}

The second model takes account of the time remaining in a game when estimating probability of a home win.  This model interacts the score difference $X_{ij}$ with a smooth function $w_\ell(t)$ of the game time,\footnote{We define this model in R with the formula {\tt home\_win $\sim$ score\_diff * poly(game\_min, degree)}. The fitted model includes a polynomial in the game time; none of the estimates $\hat\beta_1,\ldots,\hat\beta_\ell$  is statistically significant. }
\begin{eqnarray}
  \hat{p}_{ij}^{(2)} &=& g\left( \sum_{k=0}^\ell \beta_k \,t_i^k +  \hat{w}_\ell(t_i) X_{ij} \right) \cr
                            &\approx&  g\left( \hat\beta_0 + \hat{w}_\ell(t_i) X_{ij} \right)  
\label{eq:p2}
\end{eqnarray}
where $w_\ell(t)$ is a $\ell$th degree polynomial in the game time,
\ble
  \hat{w}_\ell(t) = \hat\gamma_0 + \hat\gamma_1 t + \cdots + \hat\gamma_\ell t^\ell
\qle{gamma}
The weight function $w_\ell(t)$ allows the importance of the score difference to evolve over a game.  We experimented with several choices for the degree of the polynomial and settled on $\ell=7$.  Rather than list the coefficients, Figure \ref{fi:w_of_t} graphs $w_7(t)$.  Additional points in the figure show the estimated slopes of cross-sectional logistic regressions fit to the training games at a single point in time, roughly every six minutes.  For example, the second point from the left at 6 minutes into a game is the estimated slope (0.092) of $X_{i=24,j}$ in a simple logistic regression.  At this point in the game, the slope is much smaller than that of the first model.  The polynomial $w_7(t)$  smooths the estimates and steadily grows as the game time increases.  As typical for polynomials, $\hat{w}_t(t)$ appears too volatile at the boundaries. The fit of the polynomial was not constrained but nonetheless gives minimal weight to the initial score differences. This model estimates the initial probability of a home win to be 0.589, closer to the overall proportion of home wins than estimated by the first model. The home team won 58.5\% (se 0.8) of games in the training sample, coincidentally very close to the value 58.6\% reported in \citet{mosteller92}  for home teams when averaged over four sports.

\begin{figure}
\caption{\it The left panel shows score differences and estimated probabilities $\hat{p}_{t,j}$ for three models: a simple logistic regression (dashed, equation \ref{eq:p1}), a polynomial-weighted logistic regression (solid line, equation \ref{eq:p2}), and a  martingale filter derived from the polynomial predictions (red).  The cumulative volatilities of the estimated probabilities (right) shows the reduction achieved by the martingale filter.}
\label{fi:bb_fit}
   \centerline{ \includegraphics[width=7in, height=3in]{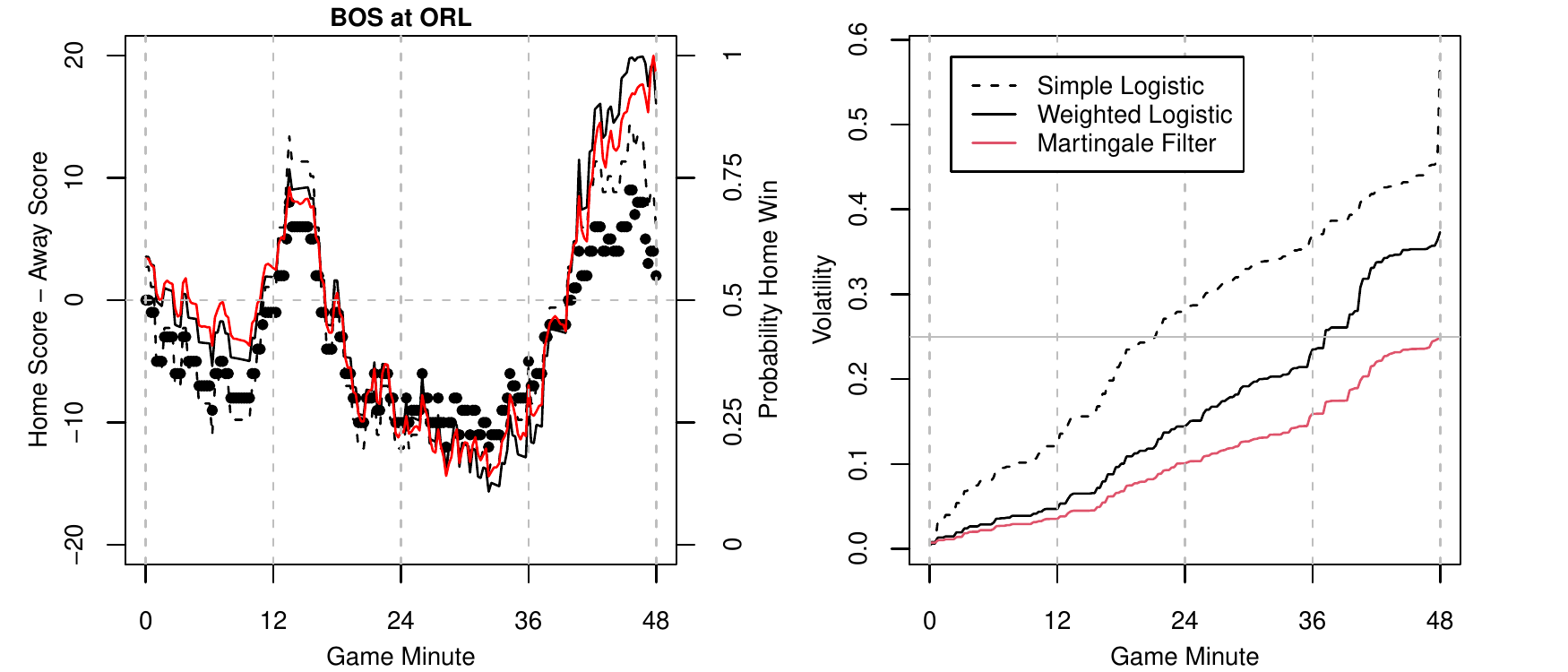} }
\end{figure}

Figure \ref{fi:bb_fit} (left panel) shows the score differences and estimated probabilities of a home win $\hat{p}^{(1)}$ and $\hat{p}^{(2)}$ for one game.  This game from the test set was played in January, 2019, with Boston visiting Orlando.  The simple logistic fit clearly places too much emphasis on the score difference early in the game and not enough late in the game.  The polynomial weights $\hat{w}_7(t)$ correct much of this weakness, but leave substantial volatility in the predictions.  The figure includes a third predictor, $\hat{p}^{(3)}$, a martingale filter described in the Appendix, that adjusts predictions in a manner that reduces volatility while improving expected accuracy.  The martingale filter reduces the MSE of  $\hat{p}^{(2)}$ even though it occasionally produces predictions outside the range [0,1], as hinted near the end of this game.  The procedure uses only the information in $\hat{p}^{(2)}$ to generate new predictions.  The right panel Figure \ref{fi:bb_fit} shows the accumulating volatility of the probability estimates. Note that every estimated probability sequence begins $\hat{p}_{-1,j} = 0.5$ and ends with the 0/1 outcome indicating if the home team won the game. The resulting accumulating volatility (compare to equation \ref{eq:st})
\ble
   S_t(\{p_{ij}\}) = \sum_{i=0}^{I_t} (p_{ij}-p_{i-1,j})^2, \quad \mbox{ where } \quad I_t = \max\{i: t_i \le t\} \;
\qle{sit}
that appears in the right panel of the figure ends far above the target level $0.5^2$ for this game for both logistic regression models. The volatility of the estimator produced by the martingale filter ends very near the expected total 0.25 for this game.  

\begin{figure}
\caption{\it The histogram on the left shows the differences in MSE between the polynomial probability $\hat{p}^{(2)}$ and the martingale filter $\hat{p}^{(3)}$ for games in the test set. The improvement is significant ($z=5.5$). Boxplots show the reduction in volatility produced first by taking account of the game time and then performing the martingale filter.}
\label{fi:bb_hist}
   \centerline{ \includegraphics[width=7in, height=3in]{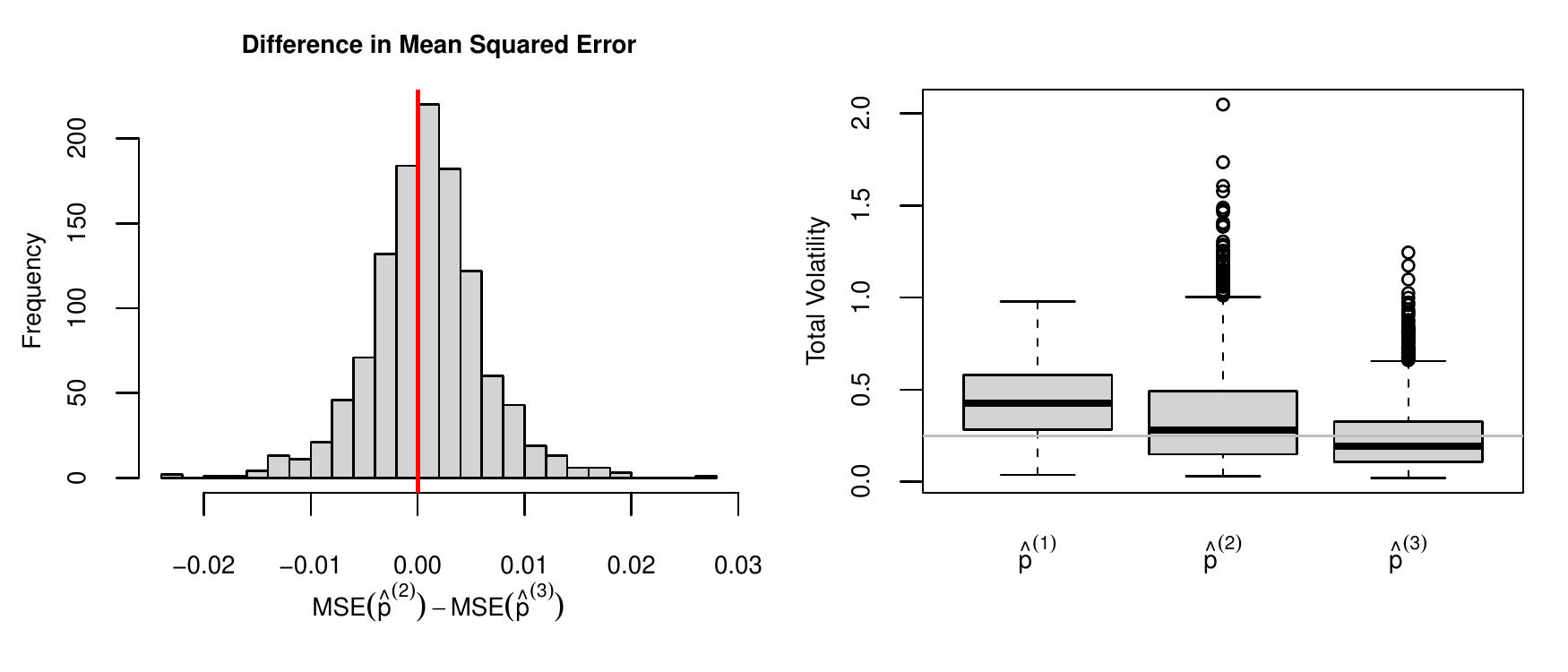} }
\end{figure}

The combination of lower volatility and lower error for the corrected predictor is typical for the games in the test set.  Figure \ref{fi:bb_hist} compares the performance of the logistic regression models and the martingale filter in the test set.  On the left, the histogram compares the MSE of the martingale filter to the polynomial-weighted logistic regression. The distribution is ever so slightly shifted to the right. The difference in average MSE is only 0.0009, but with standard error 0.00015, the shift is highly statistically significant ($z \approx 5.5$).  The improvement to the volatility summarized by the boxplots on the right of Figure \ref{fi:bb_hist} is more notable.  The probability estimates from the simple logistic model are far more volatile than appropriate for a martingale sequence.  Though better, the estimates from the polynomial weighted model also show excess volatility.  The martingale filter  fares much better, with average volatility in the test set that is not significantly different from the target 0.25; the  95\% z-interval is [0.243,  0.254].  It is interesting to note that the derivation in the appendix of the martingale filter proves that it improves MSE in expectation, but does not offer a proof of this reduction in volatility.  Table \ref{ta:summ} provides a numerical summary of the MSE and volatilies for games in the training and test sets. The MSE for predictions is accumulated over all predictions during a game.  For example, in the test set of 1161 games,
\ble
     MSE(\hat{p}^{(1)}) = \sum_{j=1}^{1161} \sum_{i=0}^{192} (Y_j - \hat{p}_{ij}^{(1)})^2/(1161 \times 193)  \;.
\qle{mse}
Though smaller in expectation, the MSE of the martingale filter is not uniformly smaller than that of the source predictor; \ie, in some games the original predictor has smaller MSE.

\begin{table}
\caption{\it Average mean squared error and total volatility of in-game predictions of the winner for games in the training and test sets. The reported MSE is the average MSE over games; see \eqn{mse}.  Shown differences in MSE and volatility are statistically significant in a paired comparison. }
\label{ta:summ}
\begin{center}
\begin{tabular}{|cc|c|c|c|} \hline
Set    &   Count  &  Predictor                   &     Average MSE     &  Average Total Volatility \cr \hline
         &               & Simple logistic           &         0.1660            &   0.428  \cr
Train &  3461     & Polynomial weighted &         0.1611            &   0.367   \cr
         &               & Martingale filter          &        0.1603             &  0.247    \cr \hline
         &               & Simple logistic           &         0.1649            &  0.437    \cr
Test  & 1161      & Polynomial weighted  &         0.1598            & 0.377    \cr
        &               & Martingale filter          &          0.1589            & 0.253   \cr  \hline
\end{tabular}
\end{center}
\end{table}

\section{ Discussion }

Some things that would be useful to include or explore are:
\begin{description}
  \item[Information flows] The rate at which the volatility accumulates measures information flows that are useful for demand planning. Consider long-lead ordering, such as required when places orders for fashion items many months ahead of the season. If the forecast volatility remains small until, say, 1 month before the FTD, then there would be little gain from ordering 1 month out rather than 6 months out -- and thus gaining a pricing advantage.   
  \item[Vector case] The analysis here treats the univariate martingale defined by a single threshold. One could combine several thresholds as shown in Figure \ref{fi:arflow} to monitor the shape of the forecast distribution more thoroughly rather than at a single threshold.
  \item[Statistical significance] Testing for excess volatility isstraightforward given one observes a collection of independent realizations, as illustrated here with simulations and basketball games.  Many other applications, however, provide highly dependent realizations, as in the case of demand forecasting. This paper does not address the question of statistical significance when the realizations are dependent.  We will address that elsewhere.
\end{description}

\section{ Appendix: Martingale Filter }

The following procedure which we call a martingale filter reduces forecast volatility while improving the accuracy of prediction. One can apply this filter to any collection of forecasts, not just those interpreted as a probability, so we switch notation for this appendix and write a sequence of forecasts for a future event as the vector $\yhat$ with elements $\yhat_t$,
\be
    \yhat = (\yhat_1, \,\ldots,\, \yhat_T)^\top \;.
\qe
We make no further assumptions about the method used to produce these forecasts. The following description of the filter illustrates  the method as applied in the basketball example in Section \ref{sec:bb}, but is not fully general and is more expository than rigorous.  Given the success in this application, we plan to develop the method further and report results in a separate paper.

We proceed by constructing a martingale representation of the predictors.  We decompose $\yhat_t$ as the sum 
\ble
   \yhat_t = M_{t|t} + M_{t|t-1} + \cdots + M_{t|0} = \sum_{s=0}^t M_{t|s} \;,
\qle{yhatt}
where the summands $M_{t|s}$ satisfy
\ble
  M_{t|s} \in {\cal F}_s, \mbox{ with } M_{t|0} \in \R \mbox{ and }\ev(M_{t|s} \given {\cal F}_{s-1}) = 0 \mbox{ for } s=1,2,\ldots,t   \;.
\qle{Mts}
In our example that predicts the chance of winning a basketball game, the first subscript $t$ in $M_{t|s}$ before the bar denotes the ``game time,'' and the second subscript $s$ denotes the conditioning information (sigma field). We treat time as discrete, such as by sampling the score every 15 seconds as done in the example.

A simplifying assumption provides an explicit construction.  Assume that we can represent each sigma field ${\cal F}_t$ as the union independent random variables ${\cal F}_t = \{Q_1, \ldots, Q_t\}$ where $Q_s$ is uncorrelated with $Q_t$ for $s \ne t$ with unit variance $\Var(Q_t) = 1$.\footnote{Independence would be too strong and uncorrelated isn't strong enough; we essentially need martingale differences but only use uncorrelated for this application.}  The collection $\{Q_1, \ldots, Q_t\}$ thus forms an orthonormal basis for ${\cal F}_t$.  To exploit this basis, note that the representation \eqn{yhatt} defines a triangular array:
\ble
  \begin{array}{ccccccc}
     \yhat_1 & = & M_{1|1} & & & & \cr
     \yhat_2 & = & M_{2|1} & + & M_{2|2} & & \cr
     \yhat_3 & = & M_{3|1} & + & M_{3|2} & + & M_{3|3} \cr
                  & \vdots& & & & & 
  \end{array}
\qle{array1}
The assumed basis allows us to write $M_{t|s} = r_{ts} Q_s $, so that each forecast $\yhat_t$ is a weighted mixture of the basis elements $Q_1,\,\ldots Q_t$:
\ble
  \begin{array}{ccccccc}
    \yhat_1  & =  &r_{11}  Q_1 & & & & \cr
    \yhat_2  & =  & r_{21} Q_1  & + &  r_{22} Q_2& & \cr
    \yhat_3  & =  & r_{31} Q_1 & + &  r_{32} Q_2& + &   r_{33}  Q_3\cr
		& \vdots  & \cr
  \end{array}
\qle{array2}
We have what amounts to a Gram-Schmidt decomposition of the vector of predictors
\ble
     \yhat =  R \, Q \;,
\qle{qr}
where $R$ is a $T\times T$ lower triangular matrix with elements $r_{ts}, s \le t$ and $Q$ is a random vector with elements $ Q = (Q_1,\ldots,Q_T)^\top$.  For the basketball application, we have $n$ independent observations of the random vectors $\yhat$ and $Q$.  We arrange these observations as rows of an $n \times T$ matrix, and we have what amounts to a QR decomposition of the matrix of predictors:
\ble
     [\yhat_1, \ldots, \yhat_T] = [Q_1, \ldots, Q_T] [ r_{st} ]_{s \le t} =  Q \, R^\top \;.
\qle{qr}

The martingale filter alters the weights of this decomposition.  The filter enforces a consistent weighting on the information in each basis element (or more generally, the information in each sigma field).  It does this by replacing each column of $R$ by the average of the nonzero elements in that column, $\ol{r}_s = \sum_{t=s}^T r_{ts}/(T-s+1)$.  Returning to the triangular equations in \eqn{array2}, we average the elements in the columns as suggested here:
\ble
  \begin{array}{cccccccc}
                    & r_{11}  Q_1 & & & & & & \cr
                    & r_{21} Q_1 &  & r_{22} Q_2 & & & & \cr
                    & r_{31} Q_1 &  & r_{32} Q_2 &  &  r_{33}  Q_3& & \cr
                    &  \vdots & & & & & & \cr
                    & r_{T1}  Q_1 &  &  r_{T2} Q_2 &  &  r_{T3} Q_3 & \cdots  & r_{TT} Q_T  \cr \hline
 \mbox{avg} & \ol{r}_1 Q_1 &   &  \ol{r}_{2} Q_2&   &  \ol{r}_3 Q_3& \cdots & \ol{r}_{T} Q_T 
  \end{array}
\qle{array3}
Compared to the input predictions laid out in \eqn{array2}, we form a martingale by accumulating the sums,\footnote{We are tempted to denote these predictors by $\ol{Y}_t$, but that symbol is too linked to the average of $Y$.}
\ble
  \begin{array}{ccccccc}
    \ytilde_1  & = & \ol{r}_1  Q_1 & & & & \cr
    \ytilde_2  & = & \ol{r}_1  Q_1 & + &  \ol{r}_2 Q_2& & \cr
    \ytilde_3  & = & \ol{r}_1  Q_1 & + &  \ol{r}_2 Q_2& + & \ol{r}_{3}  Q_3\cr
		& \vdots  & \cr
   \left[ \right. \ytilde_t  & = & \ol{M}_1 & + & \ol{M}_2 & +\cdots+ & \ol{M}_t \left. \right] \cr
		& \vdots  & \cr   
  \end{array}
\qle{array4}
The final line in this expression and several that follow display the result in the general case.  Without the basis representation, $\ytilde_t = \sum_{s=0}^t \ol{M}_s$ where $\ol{M}_s = \sum_{t=s}^T M_{t|s}/(T-s+1)$.

To state our theorem, define $\norm{X}^2 = \ev(X^\top X)$ for vectors $X$, and let  $\one$ denote a column vector of 1s with length evident from the context.  The outcome being predicted is the scalar r.v. $y$ (which we write in lower case to distinguish it from the vectors).  We prove the following: {\theorem The total expected squared error of the predictor $\ytilde$ is less than or equal to that of the initial predictor $\yhat$:  $\ev\,\norm{y\one - \ytilde}^2 \le \ev\,\norm{y\one - \yhat}^2$ \;. }

\noindent
{\bf Proof.}   Let ``$\tr$'' denote the trace operator.    The risk of the initial predictor $\yhat$ is 
\begin{eqnarray}
   R(\yhat) &=& \norm{y\one - \yhat}^2   \cr
                 &=& \norm{y\one - R\, Q}^2  \cr
                 &=& \norm{y\one - \ol{R}\, Q + \ol{R}\, Q - R\, Q}^2  \cr
                 &=& R(\ytilde) + \norm{ (\ol{R} - R) Q}^2 + 
                               2 \ev\left( (y\one - \ol{R}\, Q)^\top  (\ol{R} - R) Q \right) \cr
                 &\left[ = \right.&\left.  R(\ytilde) + \norm{ (\ol{M} - M)}^2 + 
                               2 \ev\left( (y\one - \ol{M}\, Q)^\top  (\ol{M} - M) \right) \right]\;.
\label{eq:proof}
\end{eqnarray}
The second summand measures the variability of the weights around the means of the columns of $R$:
\begin{eqnarray*}
   \norm{ (\ol{R} - R) Q}^2 
   &=& \ev \; \tr \left( Q^\top (\ol{R} - R)^\top (\ol{R} - R) Q \right) \cr
   &=& \tr  (\ol{R} - R)^\top (\ol{R} - R)  \cr
   &=& \sum_{s,t} (r_{ts} - \ol{r}_s)^2 \cr
   &\left[ = \right.&\left. \sum_{s,t} \ev\,(M_{t|s} - \ol{M}_s)^2  \right]\;.
\end{eqnarray*}
The cross product term in \eqn{proof} is zero. We show this in two steps.   First, notice that
\be
  \ev (y \one^\top (\ol{R} - R) Q) = \one^\top (\ol{R} - R) \ev(y\,Q) = 0 
        \quad \left[ = \one^\top \ev \, (\ol{M}-M)y \right]
\qe
because the sum of the deviations from the means in each column of $\ol{R} - R$ is zero. For the other part of the cross-product,
\begin{eqnarray*}
  \ev (Q^\top \ol{R}^\top  (\ol{R} - R) Q) 
   &=& \ev \; \tr (Q^\top \ol{R}^\top  (\ol{R} - R) Q) \cr
   &=& \tr \, \ol{R}^\top (\ol{R} - R) \ev(Q Q^\top)  \cr
   &=& \tr \, \ol{R}^\top (\ol{R} - R) \cr
  &\left[ = \right.&\left.  \ev\, \left( \ol{M}^\top (\ol{M}-M) \right) \right] \;.
\end{eqnarray*}
The diagonal element in position $s = 1,\ldots,T,$ of this product is zero, again because the deviations around the means sum to zero,
\be
   [\ol{R}^\top (\ol{R} - R)]_{ss} = \ol{r}_s \sum_{t=s}^T  (r_{ts}-\ol{r}_s) = 0
\qe
\QED

\bibliography{/Users/robstine/references/stat}

\begin{thebibliography}{11}
\expandafter\ifx\csname natexlab\endcsname\relax\def\natexlab#1{#1}\fi
\expandafter\ifx\csname url\endcsname\relax
  \def\url#1{\texttt{#1}}\fi
\expandafter\ifx\csname urlprefix\endcsname\relax\def\urlprefix{URL }\fi

\bibitem[{Cooper et~al.(1992)Cooper, Deneve and Mosteller}]{mosteller92}
Cooper, H., Deneve, K.~M. and Mosteller, F. (1992) Predicting professional
  sports game outcomes from intermediate scores.
\newblock \textit{Chance}, \textbf{5}, 18--22.

\bibitem[{Dawid(1984)}]{dawid84}
Dawid, A.~P. (1984) Present position and positional developments: some personal
  views, statistical theory, the prequential approach.
\newblock \textit{\JRSSA}, \textbf{147}, 278--292.

\bibitem[{Dawid(1992)}]{dawid92}
--- (1992) Prequential analysis, stochastic complexity and bayesian inference.
\newblock In \textit{Bayesian Statistics 4} (eds. J.~M. Bernardo, J.~O. Berger,
  A.~P. Dawid and A.~F.~M. Smith), 109--125. Oxford: Oxford University Press.

\bibitem[{Doob(1953)}]{doob53}
Doob, J.~L. (1953) \textit{Stochastic Processes}.
\newblock New York: Wiley.

\bibitem[{Foster and Stine(2004)}]{fosterstine04}
Foster, D.~P. and Stine, R.~A. (2004) Variable selection in data mining:
  Building a predictive model for bankruptcy.
\newblock \textit{Journal of the American Statistical Association},
  \textbf{99}, 303--313.

\bibitem[{Foster and Vohra(1998)}]{fostervohra96}
Foster, D.~P. and Vohra, R.~V. (1998) Asymptotic calibration.
\newblock \textit{Biometrika}, \textbf{85}, 379--390.

\bibitem[{Heath and Jackson(1994)}]{heath94}
Heath, D.~C. and Jackson, P.~L. (1994) Modeling the evolution of demand
  forecasts with application to safety stock analysis in
  production/distribution systems.
\newblock \textit{IIE Transactions}, \textbf{26}, 17--30.

\bibitem[{Shi and Song(2019)}]{shisong19}
Shi, J. and Song, K. (2019) A discrete-time and finite-state markov chain based
  in-play prediction model for nba basketball matches.
\newblock \textit{Communications in Statistics - Simulation and Computation},
  \textbf{48}, 1--9.

\bibitem[{Song et~al.(2020)Song, Zou and Shi}]{shisong20}
Song, K., Zou, Q. and Shi, J. (2020) Modelling the scores and performance
  statistics of nba basketball games.
\newblock \textit{Communications in Statistics - Simulation and Computation},
  \textbf{49}, 2604--2616.

\bibitem[{Stern(1994)}]{stern94}
Stern, H.~S. (1994) A {B}rownian motion model for the progress of sports
  scores.
\newblock \textit{\JASA}, \textbf{89}, 1128--1134.

\bibitem[{Toktay and Wein(2001)}]{toktay01}
Toktay, L.~B. and Wein, L.~M. (2001) Analysis of a
  forecasting-production-inventory system with stationary demand.
\newblock \textit{Management Science}, \textbf{47}, 1268--1281.

\end{thebibliography}
\bibliographystyle{/Users/robstine/work/papers/bst/rss}

\end{document}